\documentclass{iopjournal}
\usepackage{cite}
\usepackage{placeins}
\usepackage{soul}
\usepackage{xcolor}
\renewcommand{\hl}[1]{#1}

\newcommand{\hlIII}[1]{#1}
\usepackage{amsmath,amssymb,amsfonts}
\usepackage{algorithm,algorithmic}
\usepackage{hyperref}
\hypersetup{hidelinks=true}
\usepackage{textcomp}
\usepackage{booktabs}
\usepackage{graphicx}  % \resizebox in the ablation table
\usepackage{tikz}
\usetikzlibrary{decorations.pathreplacing}
\usepackage{verbatim}
\usepackage{amssymb}

\begin{document}

\articletype{Paper} %	 e.g. Paper, Letter, Topical Review...

\title{DiffEEG: A Self-Supervised Denoising Diffusion Model for Learning EEG seizure-related 
Representations}

\author{Abdulkader Helwan$^{1,*}$\orcid{0000-0000-0000-0000}, Lina Abou-Abbas$^{1,*}$\orcid{0000-0000-0000-0000}, Hussein El Amouri$^{1}$\orcid{0000-0000-0000-0000}, Belkacem Chikhaoui$^{2}$\orcid{0000-0000-0000-0000} and Khadidja Henni$^{2}$\orcid{0000-0000-0000-0000}}

\affil{$^1$Department of Electrical and Computer Engineering, Lebanese American University, Byblos, Lebanon}

\affil{$^2$Institute of Applied Artificial Intelligence, T\'ELUQ University, Montreal, Canada}

\affil{$^*$Author to whom any correspondence should be addressed.}

\email{Abedelkader.helwan@lau.edu.lb, lina.abouabbas@lau.edu.lb}

\keywords{EEG, Deep learning, denoising diffusion, foundation model, reinforcement learning.}
\begin{abstract}
Deep learning for EEG-based seizure detection faces critical challenges: severe annotation scarcity and extreme class imbalance, where ictal events comprise less than 10\% of clinical recordings. We present DiffEEG, a 9.6M-parameter self-supervised foundation model that addresses both limitations through denoising diffusion pre-training and reinforcement learning (RL)-based fine-tuning. Pre-trained on 1.3M unlabeled segments from the Temple University Hospital Seizure Corpus (TUHSZ), DiffEEG learns generic neural representations via a 1D U-Net with multi-head self-attention. For downstream adaptation, a reinforced decision layer employs policy gradient optimization to directly maximize F1-score, prioritizing sensitivity to rare seizure events over overall accuracy. Under strict patient-wise evaluation (279 patients, Leave-One-Fold-Out), DiffEEG achieves 61\% accuracy and 59\% F1 for 4-class seizure subtyping, and 81\% accuracy with 85\% weighted F1 for binary detection, reaching 59\% seizure recall under extreme imbalance (6.7\% prevalence). \hlIII{ Pre-training draws on a single corpus and uses only its non-seizure recordings, yet the resulting representation transfers, and we evaluate it under both adaptation regimes. With the backbone frozen and only a linear classifier trained on top, DiffEEG is competitive with five established EEG foundation models (EEGDM, EEGMAMBA, BIOT, LaBram, cBraMod) across two external seizure databases, neither of which it saw in pre-training, attaining the best accuracy, weighted F1 and specificity of every model compared on the Siena Scalp EEG database, although on that corpus EEGDM and CBraMod rank seizure segments better, with higher PR-AUC and sensitivity. Under fine-tuning it attains the best macro F1, PR-AUC and sensitivity against three task-trained architectures evaluated on identical TUHSZ segments, and the highest AUROC among the foundation models surveyed on abnormality detection. DiffEEG therefore shows that diffusion pre-training on one unlabelled corpus yields representations that match foundation models pre-trained on as many as twenty corpora, and exceed them on several metrics, whether the backbone is adapted or left frozen.}
\end{abstract}

\section{Introduction}
Epilepsy is one of the most prevalent neurological disorders globally, characterized by recurrent, unprovoked electrical disturbances in the brain \cite{abou2024generative}, affecting more than 50 million individuals of all ages worldwide. Electroencephalography (EEG) remains the primary non-invasive diagnostic tool for identifying seizures due to its high temporal resolution; however, the manual interpretation of EEG recordings is a labor-intensive and expensive process that necessitates highly specialized expertise \cite{abou2024generative, roy2019eeg, schirrmeister2017deep}, and visual analysis frequently suffers from significant inter-rater variability that compromises diagnostic consistency in acute clinical settings \cite{acharya2018deep}.
To overcome these constraints, researchers have increasingly employed machine learning and deep learning to identify latent spatio-temporal patterns in multi-channel EEG signals \cite{chen2024eegformer, schirrmeister2017deep, roy2019eeg, huang2026deep, henni2024imbalance, abou2021focal, lawhern2018eegnet,abou2022eeg}. The field transitioned from handcrafted feature engineering toward automatic feature extraction through convolutional and recurrent architectures \cite{ schirrmeister2017deep, roy2019eeg, thuwajit2021eegwavenet, huang2026deep, lawhern2018eegnet}, and more recently through transformer-based models capable of capturing long-range temporal dependencies via self-attention \cite{wang2024eegpt, song2021transformer, song2022eeg, ma2023tsd}.  Multimodal strategies, such as EEG-ECG synchronization toward EEG-free detection \cite{el2025toward}, were also explored in parallel, though without resolving the fundamental limitation that generalization across patients, recording setups, and institutional protocols remained fragile, and the volume of labeled EEG required to train these models reliably remained a persistent practical barrier.
These limitations motivated a shift toward EEG foundation models (FMs), large architectures pretrained in a self-supervised or unsupervised fashion on thousands of hours of unlabeled recordings from corpora such as the Temple University Hospital (TUH) EEG Corpus \cite{obeid2016temple}, and subsequently fine-tuned on specific downstream tasks \cite{wang2024eegpt, wan2023eegformer, chen2024eegformer, yang2023biot, jiang2024labram, chen2025lcm, bettinardi2025bioserenity, wang2023brainbert, peng2023wavelet2vec, peng2022tie}. 
By learning universal brain activity representations from massive and diverse data, these approaches mitigate subject-specific variability and improve adaptability to downstream clinical tasks \cite{kuruppu2026eeg}.

BioSerenity-E1 \cite{bettinardi2025bioserenity} , a self-supervised model that combines spectral tokenization with masked prediction to learn generalized representations for seizure detection and abnormality classification. EEGFormer \cite{wan2023eegformer, chen2024eegformer}, a transformer-based foundation model that employs vector quantization to derive universal EEG features that generalize across diverse clinical paradigms. EEGPT \cite{wang2024eegpt}, a hierarchical transformer that introduces spatio-temporal representation alignment enabling efficient adaptation under low-label and few-shot conditions. Wavelet2Vec \cite{peng2023wavelet2vec}, an architecture that integrates filter bank analysis with a masked autoencoder to learn multi-grained neural representations from decomposed EEG wavelets.

\hlIII{Generative modelling has entered EEG research along two distinct lines, and it is
worth separating them because the present work belongs to the second. The first uses
generative models to \emph{synthesise} EEG, most commonly to enlarge or rebalance
training sets: adversarial approaches such as WGAN-GP have been applied to seizure data to
augment under-represented ictal classes and improve downstream detection
accuracy~\cite{abou2024generative}, and adversarial training has likewise been used to
improve robustness in seizure models~\cite{zhang2020adversarial}. The second line uses the
generative \emph{objective} as a pretext task, discarding the sampling capability and
retaining the learned encoder as a feature extractor. Denoising diffusion probabilistic
models~\cite{ho2020denoising,nichol2021improved} are well suited to this because their
training signal, predicting the noise added at a known corruption level, requires the
network to model the structure of the data at every scale. Diffusion models are now
established for one-dimensional signals, with WaveGrad~\cite{chen2021wavegrad} and
DiffWave~\cite{kong2021diffwave} demonstrating high-fidelity waveform modelling in audio,
and architectural conventions for the conditioning pathway are inherited from the image
diffusion literature~\cite{dhariwal2021diffusion,rombach2022high}.}

\hlIII{DiffEEG belongs to this second line. We train a denoising diffusion model on
unlabelled EEG, but we do not generate synthetic EEG at any stage, and none of the results
reported in this paper involve synthesised or augmented signals. The reverse process is
used solely as a self-supervised objective, after which the encoder is probed at fixed
points on the noise schedule to obtain representations for classification. Comparisons
against data-augmentation methods are therefore not directly applicable; the appropriate
comparison is against other representation-learning approaches, which is the basis of the
benchmarking in Section~\ref{sec:benchmarking}.}
\hlIII{Most directly related to the present work, EEGDM~\cite{puah2025eegdm} applies a
denoising diffusion objective to EEG for the purpose of \emph{representation extraction}
rather than signal generation, using a structured state-space backbone pre-trained on the
TUEV corpus and adapted downstream through a latent fusion transformer while the
pre-trained parameters remain frozen. The present work differs in the specific instantiation and in the
evaluation setting rather than in that premise: DiffEEG uses a 1-D U-Net with multi-head
self-attention and FiLM timestep conditioning, is pre-trained on the non-seizure portion
of TUHSZ, extracts features by probing the denoiser at multiple points along the noise
schedule, and is adapted by partial fine-tuning of the backbone together with a decision
layer optimised for the target metric. Its evaluation is directed at severe class
imbalance in continuous seizure monitoring, where ictal events form a small minority of
the recording, rather than at the balanced downstream benchmarks on which diffusion-based
EEG representations have previously been assessed. To place the two on comparable
footing we additionally benchmark DiffEEG directly against EEGDM and other foundation models under a single shared
protocol in Section~\ref{sec:external_probe}, and report the outcome in both directions.}

Despite demonstrating significant success in generalized representation learning \cite{wang2024eegpt, wan2023eegformer, chen2024eegformer, yang2023biot, jiang2024labram, chen2025lcm}, the downstream clinical evaluations of these models remain largely confined to standard and relatively balanced benchmarks such as CHB-MIT \cite{PhysioNet-chbmit-1.0.0} and the Temple University Abnormal and EEG Events corpora \cite{obeid2016temple}, which do not adequately capture the severe class imbalances inherent to acute continuous monitoring. Consequently, existing architectures such as BIOT \cite{yang2023biot}, LaBraM \cite{jiang2024labram}, and LCM \cite{chen2025lcm} are rarely optimized for or evaluated on the TUHSZ corpus for binary seizure detection or granular seizure subtype classification, where epileptiform events constitute a minor fraction of the overall recording \cite{obeid2016temple}. 

A critical gap therefore remains in the literature regarding the application of foundation models to realistic clinical monitoring conditions \cite{kuruppu2026eeg}, where seizure events constitute only approximately 6.7\% of total recording time \cite{wang2024eegpt, wan2023eegformer, chen2024eegformer, yang2023biot, jiang2024labram, chen2025lcm}, and where many current models require significant computational resources that limit their utility in resource-constrained medical environments \cite{wang2024eegpt,bettinardi2025bioserenity}.

To address these challenges, we propose DiffEEG, a 9.6M-parameter self-supervised foundation model tailored for seizure detection in continuous monitoring and other EEG-related tasks. DiffEEG is pretrained on 1.3M unlabeled segments from the TUHSZ corpus using a diffusion-based objective \cite{ho2020denoising} applied to a 1D U-Net with multi-head self-attention, learning robust neural representations without requiring any annotations. A reinforced decision layer (RDL) subsequently employs policy gradient (PL) optimization to directly maximize the F1-score, sharpening sensitivity to rare seizure events rather than optimizing for overall accuracy. Unlike previous architectures, DiffEEG is explicitly fine-tuned and benchmarked on the highly imbalanced TUHSZ dataset and on complex seizure subtyping tasks, demonstrating that foundational representations can be strictly optimized for real-world clinical environments where maximizing sensitivity to rare events is paramount. The main contributions of this work are:
\begin{enumerate}
\item A diffusion-driven pretraining strategy that captures transferable neural representations from non-seizure EEG.
\item A reinforcement-based adaptation mechanism that tunes the decision boundary toward rare seizure events, whose contribution we isolate and report to be a shift of the operating point rather than a gain in discrimination.
\item Demonstration of label efficiency and scalability through linear probing, few-shot learning, and partial fine-tuning.
\item A transparent and reproducible framework for clinical translation. GitHub: https://github.com/AbdulkaderHelwan/Diff-EEG

\end{enumerate}

\section{Methods}
\label{sec:methods}
DiffEEG is organized as a two-stage framework. In the first stage, a 1D U-Net encoder built from residual blocks with FiLM conditioning, multi-head self-attention, and sinusoidal position embeddings is pretrained in a fully self-supervised manner through a denoising diffusion objective \cite{ho2020denoising}, learning to recover clean EEG segments from corrupted inputs without any annotation. In the second stage, the pretrained encoder is adapted to seizure classification through a reinforced decision layer that employs policy gradient optimization to directly maximize the F1-score, prioritizing sensitivity to rare ictal events over global accuracy. 
\begin{figure}[t]
\centering
\includegraphics[width=0.5\textwidth]{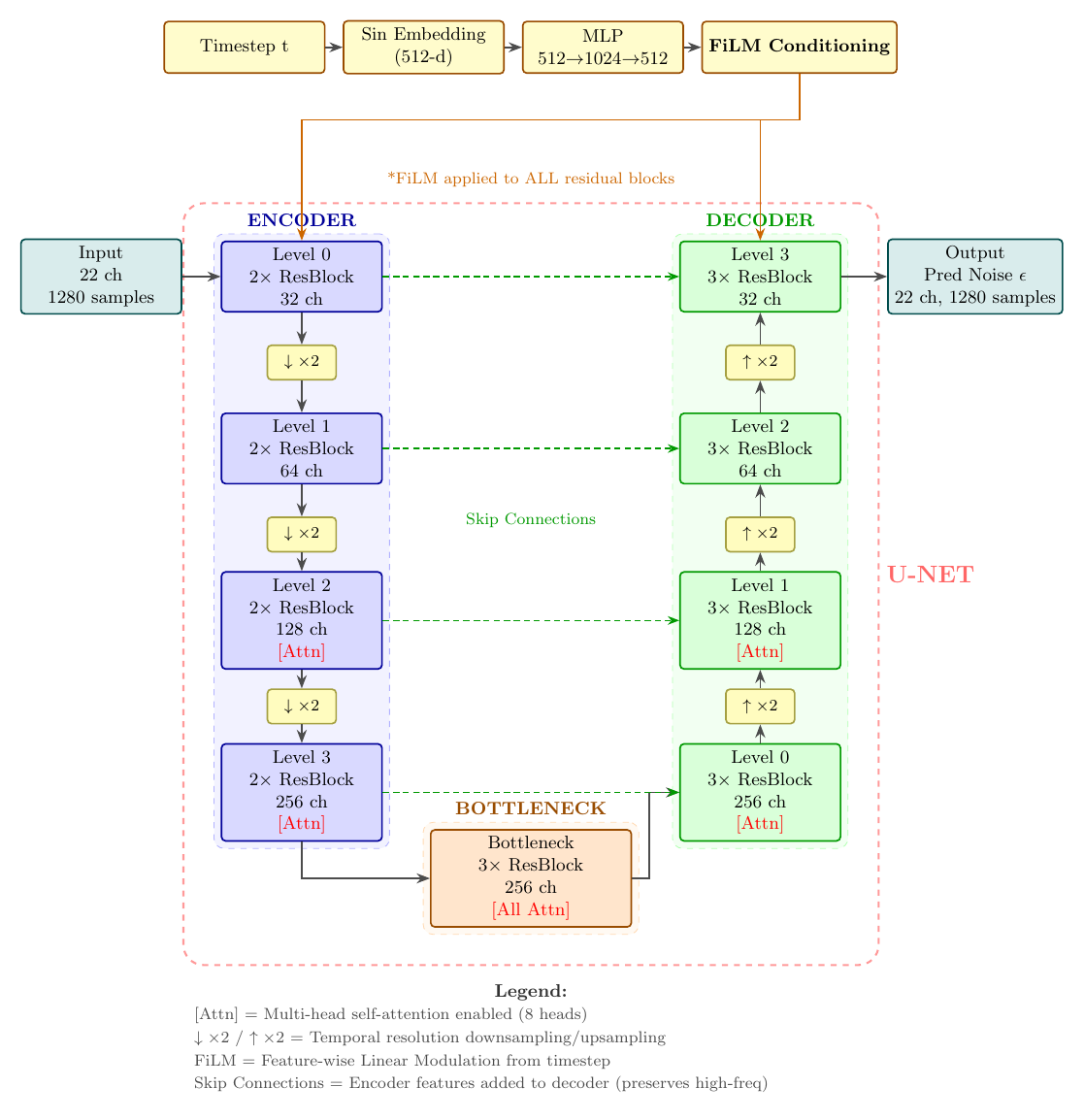}
\caption{ Overview of the DiffEEG 1D U-Net architecture, showing the encoder, bottleneck, and decoders pathway used for feature extraction.}
\label{fig:architecture}
\end{figure}

\subsection{Diffusion Process Formulation}
The diffusion process follows the framework established by Ho et al. \cite{ho2020denoising}, comprising a forward noising process and a learned reverse denoising process. Given an original EEG signal $x_0 \in \mathbb{R}^{C \times L}$, where $C$ represents the number of EEG channels and $L$ denotes the temporal length, the forward process progressively adds Gaussian noise across $T$ timesteps according to:
\begin{equation}
x_t = \sqrt{\bar{\alpha}_t} x_0 + \sqrt{1 - \bar{\alpha}_t} \epsilon, \quad \epsilon \sim \mathcal{N}(0, \mathbf{I})
\label{eq:forward_diffusion}
\end{equation}
where $\bar{\alpha}_t = \prod_{i=1}^t \alpha_i$ and $\alpha_i = 1 - \beta_i$. The variance schedule $\beta_t$ is computed using a cosine-based approach following Nichol and Dhariwal \cite{nichol2021improved}:
\begin{equation}
\beta_t = \text{clip}\left(1 - \frac{\bar{\alpha}_{t+1}}{\bar{\alpha}_t}, 0, 0.999\right)
\label{eq:beta_schedule}
\end{equation}
where $\bar{\alpha}_t$ is derived from:
\begin{equation}
\bar{\alpha}_t = \frac{\cos^2\left(\frac{t/T + s}{1 + s} \cdot \frac{\pi}{2}\right)}{\cos^2\left(\frac{s}{1 + s} \cdot \frac{\pi}{2}\right)}
\label{eq:cosine_alpha}
\end{equation}
with $s = 0.008$ as a small offset to prevent the schedule from reaching zero too early. The cosine schedule provides more stable training dynamics compared to linear alternatives, particularly important for the high-frequency characteristics of EEG signals.

\hl{\textit{Noise Schedule and Data Conditioning:} EEG is spectrally coloured, with an approximately $1/f$ power spectrum. The Gaussian forward process is not a model of the noise already present in the recording; it is a designed corruption path from the data distribution to $\mathcal{N}(0,\mathbf{I})$, and it makes no assumption that the signal being corrupted is itself white. The distinction matters for how the model learns. Because the injected perturbation $\epsilon \sim \mathcal{N}(0,\mathbf{I})$ is spectrally flat, it is statistically distinguishable from the strongly non-flat spectrum of the underlying EEG; all $1/f$ and rhythmic structure resides in $x_0$, so the denoiser can only predict $\epsilon$ by learning that structure. Adopting a coloured forward process instead would make the corruption spectrally similar to the signal and would remove much of that separability, in addition to forfeiting the closed-form marginal $q(x_t \mid x_0)$ and the reparameterization trick that make single-step training tractable. The cosine variance schedule then controls only \textit{how much} noise is added at each timestep, not its spectral shape: early timesteps retain most of the signal so the denoiser must resolve fine morphology, while later timesteps dominate the signal so it must rely on global structure. Preprocessing supports this separation: the 1--75\,Hz bandpass and 50\,Hz notch remove the sub-1\,Hz drift that carries the steepest part of the $1/f$ slope, and channel-wise normalization equalizes scale across electrodes, though we note these steps attenuate rather than eliminate the spectral tilt.}
The reverse process, which the neural network learns to approximate, is given by:
\begin{equation}
p_\theta(x_{t-1} | x_t) = \mathcal{N}\left(\mu_\theta(x_t, t), \sigma_t^2 \mathbf{I}\right)
\label{eq:reverse_process}
\end{equation}
where the mean $\mu_\theta(x_t, t)$ is computed from the network's noise prediction $\epsilon_\theta(x_t, t)$.
\subsection{1D U-Net Architecture Overview}
\label{sec:network}
The core neural network $\epsilon_\theta(x_t, t)$ employs a U-Net-based encoder-decoder architecture augmented with multi-head self-attention mechanisms and feature-wise linear modulation (FiLM) conditioning \cite{perez2018film}. This architectural configuration enables the model to capture both local temporal patterns characteristic of EEG signals and long-range dependencies across channels and time.
The network receives as input the noisy EEG sample $x_t \in \mathbb{R}^{C \times L}$ concatenated with a timestep embedding, producing as output the predicted noise component $\epsilon_\theta(x_t, t) \in \mathbb{R}^{C \times L}$. The architecture follows an encoder-bottleneck-decoder structure with symmetric skip connections preserving high-frequency information essential for accurate EEG signal reconstruction. Figure 1 shows the full conditional denoising diffusion model, showing the U-Net blocks, sinusoidal position embeddings, and FiLM conditioning.  
\begin{enumerate}
\item \textbf{Encoder }
The encoder pathway consists of multiple resolution stages, each containing residual blocks with progressive channel expansion. Given base model channels $m = 32$ and channel multipliers $[1, 2, 4, 8]$, the encoder produces feature maps with dimensions $[m, 2m, 4m, 8m]$ across successive levels. At each resolution level $i$, the output channels are computed as $C_i = m \cdot \text{multipliers}[i]$.

\hl{The use of channel multipliers [1, 2, 4, 8] enables progressive feature abstraction through the encoder, allowing hierarchical multi-scale representations of EEG signals. Rather than imposing rigid dimensionality reduction, this design allows the network to learn increasingly abstract temporal and spectral patterns at deeper encoder levels, while early layers preserve fine-grained signal details critical for EEG analysis. As we progress from 32 to 256 channels across encoder levels, the receptive field size grows exponentially, enabling simultaneous capture of local discriminative features (e.g., spike morphology at millisecond scales) and long-range dependencies (e.g., seizure evolution over multiple seconds).}

Each level contains $N = 2$ residual blocks processing feature maps at that resolution before downsampling. \hl{Downsampling via strided convolution (kernel size 3, stride 2) is essential for EEG analysis: it reduces temporal resolution by half while enabling exponential growth of receptive fields, allowing the model to capture long-range temporal dependencies spanning multiple seconds, which is critical for detecting seizures that unfold dynamically over time.} Downsampling employs strided convolution with kernel size 3 and stride 2, reducing temporal resolution by half while preserving channel count.
\item \textbf{Bottleneck}
The bottleneck processes the most compressed representation, consisting of three stacked residual blocks with self-attention enabled. The bottleneck operates at the finest temporal compression $(L/8)$ with the highest channel count $(8m = 256)$, enabling global context aggregation before reconstruction.

\hl{\textit{Bottleneck Design Rationale:} The bottleneck configuration (3 residual blocks, temporal compression $L/8$, channel count $8m = 256$) follows directly from the encoder specification rather than from independent tuning. The compression factor $L/8$ is not a free parameter: it is produced by the three stride-2 downsampling stages implied by the four-level multiplier schedule $[1,2,4,8]$, and the channel count $8m = 256$ is likewise the final entry of that schedule. At this resolution a 5-second segment is reduced to 160 time steps, which is short enough for the quadratic cost of self-attention to be negligible while retaining the temporal structure needed for global context aggregation. Three residual blocks were used so that the bottleneck contains more than a single attention stage while keeping activation memory within the budget of the single A100 used for pretraining. We did not sweep bottleneck depth: because the bottleneck is part of the pretrained backbone, each additional configuration would require repeating the full pretraining run rather than a fine-tuning run, which was beyond the compute available for this study.}
\item \textbf{Decoder}
The decoder pathway mirrors the encoder structure with upsampling operations replacing downsampling. \hl{Each upsampling stage employs linear interpolation (scale factor 2) followed by a learned $3\times1$ convolution. This two-step operator is not a linear model of the signal, and the distinction is worth making explicit given that EEG is non-linear: interpolation fixes only the \textit{resampling grid}, determining where samples are placed, while the subsequent learned convolution, followed by the non-linear activations and residual blocks of that decoder level, supplies the non-linear, signal-specific reconstruction. The composition is therefore no less expressive than a learned upsampling operator of the same receptive field. We prefer it to strided transposed convolution because transposed convolutions produce uneven overlap when the kernel size is not divisible by the stride, yielding the well-documented checkerboard artefact \cite{odena2016deconvolution}; in a one-dimensional signal this appears as periodic amplitude ripple at the upsampling period, which is an undesirable failure mode for a model whose output is itself an EEG waveform.} Skip connections from corresponding encoder levels are fused via channel-wise addition after temporal interpolation to align dimensions between encoder and decoder features.
\end{enumerate}
\subsection{Residual Blocks with FiLM Conditioning}
\label{sec:residual_film}
Each residual block implements a conditioning mechanism based on FiLM, originally introduced for neural network conditioning in generative modeling \cite{perez2018film}. The FiLM layer modulates intermediate feature representations based on the diffusion timestep, enabling the network to adapt its processing based on the noise level present in the input. The residual block structure is shown in Figure 2. 
\begin{figure}[t]
\centering
% Adjust width to fit single column (usually \columnwidth) 
% or use 0.9\columnwidth for some margin
\includegraphics[width=\columnwidth]{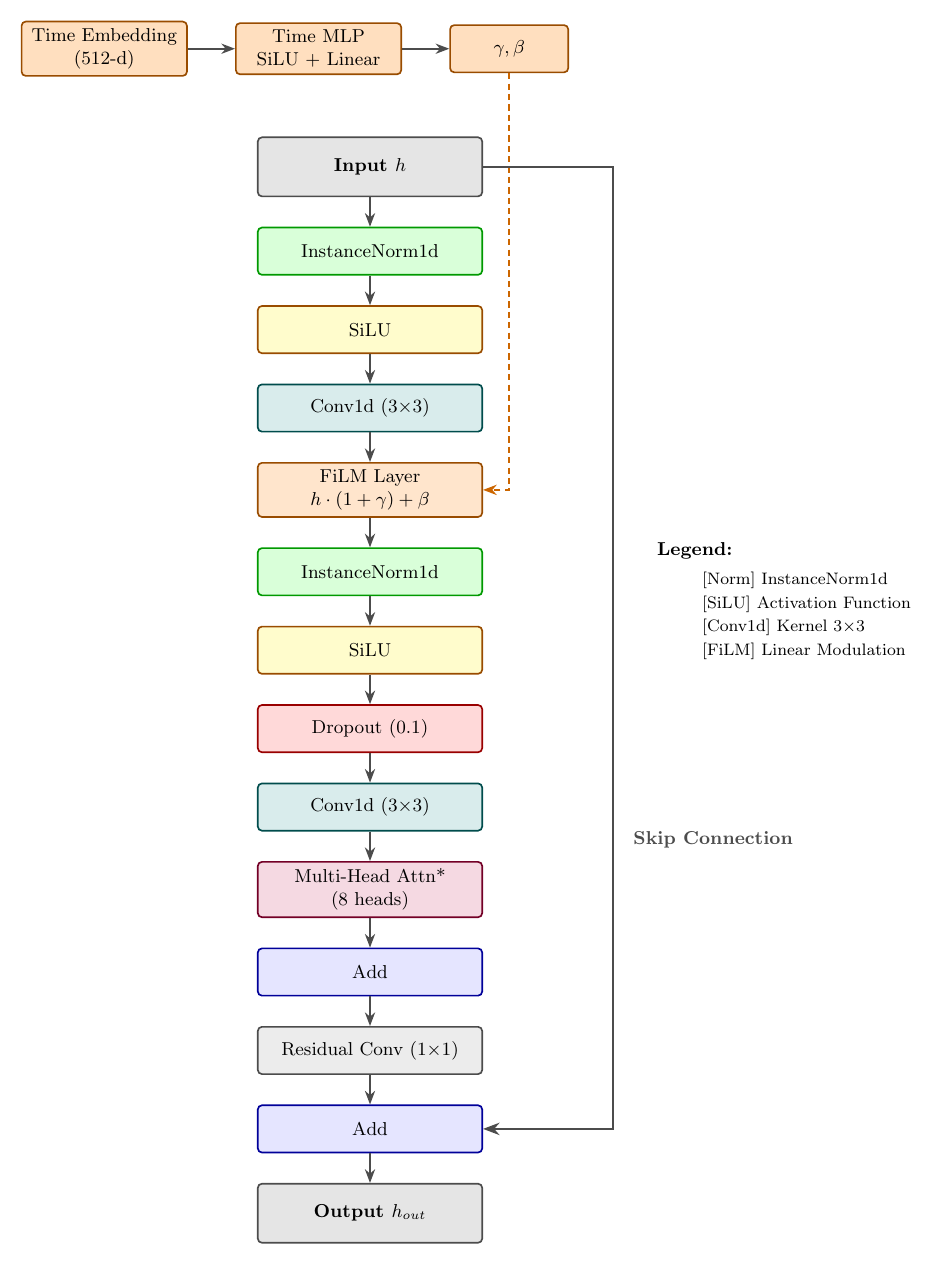}

\caption{Detailed structure of the residual block featuring Feature-wise Linear Modulation (FiLM) for timestep-based conditioning.}
\label{fig:residual_block}
\end{figure}

Given input features $h \in \mathbb{R}^{B \times C \times L}$ where $B$ denotes batch size, $C$ represents channels, and $L$ represents sequence length, the residual block computes:
\begin{enumerate}
\item \textbf{First normalization and convolution:}
\begin{equation}
h' = \text{Conv}_1(\text{SiLU}(\text{InstanceNorm}(h)))
\label{eq:first_conv}
\end{equation}
\hl{Convolutional operations employ 3$\times$1 kernels (3 temporal positions) with padding='same' to preserve sequence length and capture local temporal dependencies in EEG signals. The kernel size of 3 is selected to balance local feature extraction (capturing dependencies within $\sim$12 ms at 256 Hz sampling rate) while remaining computationally efficient. Instance Normalization stabilizes training by normalizing features per sample rather than across batch, reducing sensitivity to batch composition.}

\item \textbf{FiLM conditioning:} The timestep embedding $t_{\text{emb}} \in \mathbb{R}^{D}$ is processed through a conditioning network:
\begin{equation}
[\gamma, \beta] = W_2 \cdot \text{SiLU}(W_1 \cdot t_{\text{emb}} + b_1) + b_2
\label{eq:film_params}
\end{equation}
where $\gamma, \beta \in \mathbb{R}^{C}$ are per-channel scale and bias vectors. \hl{The conditioning pathway has two stages, which we state explicitly here:}
\begin{itemize}
\item \hl{A \textit{shared} timestep MLP, applied once per forward pass, maps the sinusoidal code $D \rightarrow 2D \rightarrow D$ with a SiLU non-linearity between the two linear layers (Eq.~\eqref{eq:time_mlp}).}
\item \hl{A \textit{per-block} projection, preceded by a SiLU activation, maps the shared code to the modulation parameters of that block, $\mathrm{Linear}(D, 2C)$, whose output is split into $\gamma$ and $\beta$.}
\end{itemize}
\hl{Each residual block therefore applies a single affine map from the shared embedding to its $2C$ modulation coefficients, rather than a deeper per-block MLP. This keeps the conditioning pathway linear in the block width $C$ and matches the parameterization used by DDPM \cite{ho2020denoising} and subsequent diffusion U-Nets; the FiLM formulation itself follows Perez et al. \cite{perez2018film}. As shown in Section~\ref{sec:position}, the embedding width is set to $D = 2\max_l C_l$ so that this projection is never rank-limiting at any level of the network.}
The FiLM operation applies:
\begin{equation}
h'' = h' \cdot (1 + \gamma) + \beta
\label{eq:film_application}
\end{equation}
\item \textbf{Second normalization, activation, dropout, and convolution:}
\begin{equation}
h''' = \text{Conv}_2(\text{SiLU}(\text{InstanceNorm}(h'')))
\label{eq:second_conv}
\end{equation}
\item \textbf{Attention computation}:
\begin{equation}
h_{\text{attn}} = \text{MultiHeadAttention}(h''')
\label{eq:attention}
\end{equation}
\begin{equation}
h''' = h''' + \text{Proj}(h_{\text{attn}})
\label{eq:attention_residual}
\end{equation}
\item \textbf{Residual connection:}
\begin{equation}
h_{\text{out}} = h''' + \text{ResidualConv}(h_{\text{in}})
\label{eq:residual_connection}
\end{equation}
\end{enumerate}
The residual connection employs a $1 \times 1$ convolution when input and output channel dimensions differ, enabling flexible channel expansion throughout the network.
\subsection{Multi-Head Self-Attention}
\label{sec:attention}
Self-attention modules are strategically positioned within the network to balance computational efficiency with representational capacity:
\begin{itemize}
\item \textbf{Encoder attention:} Attention is applied in the deeper encoder levels (second half of channel multipliers) and only at the final residual block of each level, capturing high-level abstractions while maintaining local feature processing in earlier stages.
\item \textbf{Bottleneck attention:} All three bottleneck residual blocks employ self-attention, enabling global context aggregation at the most compressed representation.
\item \textbf{Decoder attention:} Attention is applied in earlier decoder levels (first half of channel multipliers) at the first residual block of each level, facilitating reconstruction of spatial-temporal patterns from high-level representations.
\end{itemize}
This asymmetric attention placement reflects the progressive abstraction characteristic of encoder-decoder architectures: encoder attention captures hierarchical feature representations while decoder attention focuses on reconstructing detailed output patterns.
\subsection{Sinusoidal Position Embeddings}
\label{sec:position}
Temporal information within the diffusion process is encoded using sinusoidal position embeddings, following the original transformer architecture introduced by Vaswani et al. \cite{vaswani2017attention}. These fixed positional embeddings provide the model with awareness of temporal position within the diffusion trajectory without introducing learnable parameters.
\subsubsection{Embedding Computation}
For a timestep $t \in [0, T-1]$ and embedding dimension $D$, the position embedding vector $PE \in \mathbb{R}^D$ is constructed as:
\begin{equation}
PE_{(2i)} = \sin\left(\frac{t}{10000^{2i/D}}\right), \quad PE_{(2i+1)} = \cos\left(\frac{t}{10000^{2i/D}}\right)
\label{eq:sinusoidal_pe}
\end{equation}
where $i \in [0, D/2)$ indexes the dimension position. This formulation produces embeddings where different positions yield distinct vectors while maintaining a structured relationship that enables the model to generalize to unseen positions.

\hl{\textit{Embedding Dimension Selection:}} \hlIII{The embedding width $D=512$ was chosen to match the timestep-conditioning interface rather than tuned against validation performance. Each residual block conditions on $t$ through a projection that emits one scale and one shift coefficient per output channel, $\mathrm{Linear}(D, 2C) \rightarrow (\gamma, \beta)$. The widest levels of the encoder carry $C = 8m = 8 \times 32 = 256$ channels and therefore consume $2C = 512$ modulation coefficients, so setting}
\begin{equation}
D \;=\; 2\max_{l} C_l \;=\; 512
\label{eq:embed_dim}
\end{equation}
\hlIII{renders the deepest FiLM projection square ($512 \rightarrow 512$), so that the timestep code is not linearly compressed at the point where it modulates the largest number of feature maps, while any $D > 512$ would enlarge the projection matrix in every residual block without increasing its attainable rank. We state the limits of this argument rather than overstating it. Since the timestep code is a deterministic function of the scalar $t$, the set of modulation vectors the network can realise consists of at most $T$ points lying on a one-dimensional curve, whatever the value of $D$; the rank of the projection therefore does not bind in the way a fully general linear map would suggest, and we do not claim that a smaller $D$ would measurably degrade performance. The choice is also not free of cost: at $D=512$ the shared timestep MLP accounts for 1{,}050{,}112 parameters, roughly a third of everything evaluated during feature extraction. $D=512$ should accordingly be read as a convention aligned to the widest feature width, consistent with the base width $m=32$ and the multipliers $[1,2,4,8]$, and not as a quantity derived from first principles or selected by search.}

\hl{The choice is also well conditioned with respect to timestep resolution: the sinusoidal code of Eq.~\eqref{eq:sinusoidal_pe} supplies $D/2 = 256$ sine/cosine frequency pairs spaced geometrically over $[1, 10^{4}]$, which resolves the $T=1000$ discrete timesteps with a wide margin so that adjacent noise levels receive well-separated codes. It further follows established practice: diffusion U-Nets size the timestep embedding to the network's \textit{widest} feature width rather than its input-stem width, with ADM \cite{dhariwal2021diffusion} and the latent-diffusion U-Net \cite{rombach2022high} both setting it to four times the base channel count, which in those configurations coincides with the widest channel width. The injection of a sinusoidal embedding into every residual block follows DDPM \cite{ho2020denoising}, its scale/shift form follows FiLM \cite{perez2018film}, and the same mechanism is employed by one-dimensional signal diffusion models such as WaveGrad \cite{chen2021wavegrad} and DiffWave \cite{kong2021diffwave}. Our $D=512$ for a 256-channel widest level is the direct analogue of this convention.}

\hl{We note that information-theoretic model-selection criteria (AIC/BIC) are not well posed for this particular choice. Such criteria compare candidate models differing in parameter count under a common, correctly specified likelihood fitted by maximum likelihood. Here $D$ is pinned to $2\max_l C_l$ by the architecture, so varying it independently of the channel schedule would alter the conditioning mechanism itself rather than the model order. Moreover, a denoising diffusion model is trained on a variational bound with a reweighted $\epsilon$-prediction objective and is regularized implicitly by dropout, early stopping and the optimizer, so its effective degrees of freedom are not its parameter count. Validation loss and downstream task performance are therefore the established selection criteria in this setting.}

The timestep is normalized to $[0, 1]$ before computing embeddings, ensuring consistent scaling regardless of the absolute number of diffusion timesteps.

\hl{\textit{Fixed Timestep Schedule:} The diffusion process employs a fixed schedule of $T=1000$ timesteps with cosine-based variance scaling, following standard practice \cite{ho2020denoising,nichol2021improved}. A fixed schedule is not merely conventional here; it is what makes the rest of the method well defined. Single-step training relies on the closed-form marginal $q(x_t \mid x_0) = \mathcal{N}(\sqrt{\bar{\alpha}_t}x_0, (1-\bar{\alpha}_t)\mathbf{I})$, whose coefficients $\bar{\alpha}_t$ are determined by a schedule fixed in advance; making the schedule adaptive would change $\bar{\alpha}_t$ during training and dissolve the correspondence between a timestep index and a specific noise level. }\hlIII{The correspondence matters for downstream probing as well, though in a narrower sense than a corruption level shared across stages: because no noise is added at inference (Section~\ref{sec:detection}), what a fixed schedule guarantees is that index $t$ was associated with one and the same noise level throughout pre-training, so the conditioning the network learned for that index is internally consistent. It does not make the index correspond to any corruption present at inference, since there is none.} \hl{A fixed schedule additionally keeps the pretraining objective directly comparable with published diffusion models and makes runs reproducible across implementations. During training, timesteps are sampled uniformly from $\{1,\ldots,T\}$ so that the denoiser is trained at all noise levels.}
\subsubsection{Time Embedding Network}
The raw timestep embedding undergoes further processing through a multilayer perceptron before FiLM conditioning:
\begin{equation}
t_{\text{emb}} = \text{MLP}(PE(t))
\label{eq:time_mlp}
\end{equation}
The MLP comprises:
\begin{enumerate}
\item Linear projection: $D \rightarrow 2D$
\item SiLU activation
\item Linear projection: $2D \rightarrow D$
\end{enumerate}
This two-layer network transforms the sinusoidal embedding into a richer representation suitable for conditioning the residual blocks.

\section{Experiments and Setup}
\label{sec:ExperimenetsAndSetup}
To address the spatio-temporal complexity of EEG signals and the inherent scarcity of labeled seizure subtype data, we propose a multi-stage hierarchical learning framework. This approach transitions from general neural representation learning to specialized clinical classification.
\subsection{Self-Supervised Diffusion Pre-training}
The foundational stage of our framework involves pre-training a denoising EEG diffusion model on a large-scale dataset of non-seizure EEG recordings. Unlike traditional discriminative pre-training, the diffusion objective forces the model to learn the underlying manifold of non-seizure brain activity by reversing a stochastic degradation process.

We employ a 1D U-Net architecture integrated with residual blocks and multi-head self attention mechanisms. Given a clean EEG segment $x_0$, a forward diffusion process adds Gaussian noise according to a Cosine Beta Schedule over $T=1000$ timesteps, resulting in a latent representation $x_T$. The model $\epsilon_\theta$ is trained to minimize the Mean Squared Error (MSE) between the added noise and the predicted noise: 
\begin{equation}
\mathcal{L}_{diff} = \mathbb{E}_{x_0, \epsilon, t} \left[ \| \epsilon - \epsilon_\theta(x_t, t) \|^2 \right]
\end{equation}
Additionally, a reconstruction loss is incorporated to ensure the high-fidelity preservation of the signal's structural components, providing a robust initialization for subsequent tasks.

 \subsection{Binary Seizure Detection via Fine-Tuning (with RL)}
\label{sec:detection}

\hl{\textit{Detection Pipeline Overview:} Binary seizure detection follows a three-stage pipeline: (1) multi-timestep feature extraction via diffusion probing, (2) reinforced decision layer processing, and (3) deterministic threshold decision. Algorithm 1 details this complete procedure.}

\begin{algorithm}[t]
\caption{Binary Seizure Detection Pipeline}
\label{alg:seizure_detection}
\begin{algorithmic}
\STATE \textbf{Input:} \textit{clean} EEG segment $x \in \mathbb{R}^{C \times L}$ (no noise is added), pretrained encoder $\epsilon_\theta$, probe timesteps $\mathcal{T}_{\text{probe}} = \{50, 250, 500, 750, 950\}$
\STATE \textbf{Output:} Seizure probability $p_{\text{seizure}}$, binary decision $\hat{y}$
\STATE
\STATE \textbf{Stage 1: Multi-Timestep Feature Extraction}
\FOR{each $t \in \mathcal{T}_{\text{probe}}$}
\STATE $h^{(t)} \leftarrow \text{Encoder}_\theta(x, t)$ \quad \COMMENT{$t$ conditions the encoder; $x$ is not corrupted}
\STATE $\mathbf{f}_\ell^{(t)} \leftarrow \text{GlobalAveragePool}(h_\ell^{(t)})$ \quad $\forall \ell \in \{1,2,3,4\}$ \quad \COMMENT{Extract encoder level features}
\STATE $\text{feat}_t \leftarrow [\mathbf{f}_1^{(t)} \oplus \mathbf{f}_2^{(t)} \oplus \mathbf{f}_3^{(t)} \oplus \mathbf{f}_4^{(t)}]$ \quad \COMMENT{Concatenate across levels}
\STATE $\mathbf{z} \leftarrow \mathbf{z} \oplus \text{feat}_t$ \quad \COMMENT{Accumulate to feature vector}
\ENDFOR
\STATE $\mathbf{z} \in \mathbb{R}^{2400}$ \quad \COMMENT{Final accumulated features ($D_{\text{total}} = 5 \times 480$)}
\STATE
\STATE \textbf{Stage 2: Reinforced Decision Layer}
\STATE $l_{\text{seizure}}, l_{\text{normal}} \leftarrow \text{ClassificationHead}(\mathbf{z})$ \quad \COMMENT{Base logits}
\STATE $a \leftarrow \pi_\phi(\mathbf{z})$ \quad \COMMENT{RDL policy adjustment}
\STATE $\tilde{l}_{\text{seizure}} \leftarrow l_{\text{seizure}} + a$ \quad \COMMENT{Boost seizure sensitivity}
\STATE $\tilde{l}_{\text{normal}} \leftarrow l_{\text{normal}} - a$ \quad \COMMENT{Reduce false negatives}
\STATE $p_{\text{seizure}} \leftarrow \sigma(\tilde{l}_{\text{seizure}} - \tilde{l}_{\text{normal}})$ \quad \COMMENT{Adjusted probability}
\STATE
\STATE \textbf{Stage 3: Inference Decision (Deterministic Threshold)}
\IF{$p_{\text{seizure}} > 0.5$}
\STATE $\hat{y} \leftarrow \text{Seizure}$
\ELSE
\STATE $\hat{y} \leftarrow \text{Normal}$
\ENDIF
\STATE
\RETURN $(p_{\text{seizure}}, \hat{y})$
\end{algorithmic}
\end{algorithm}

\hl{Binary seizure detection employs the same multi-timestep feature extraction strategy (explained in Algorithm 1 and Appendix D.2), probing the diffusion backbone at $\mathcal{T}_{\text{probe}} = \{50, 250, 500, 750, 950\}$ and aggregating hierarchical features via global average pooling. 

The diffusion process itself is used only during pre-training. Downstream, no noise is added to the signal and no noise is predicted: the clean segment is passed to the encoder together with a timestep index, and the intermediate activations are read out. The timestep therefore acts purely as a conditioning input. Because pre-training taught the network to process a signal differently according to the noise level it was told to expect, evaluating the same clean segment at five different timesteps yields five conditioned views of it, which we concatenate. We measure in Section~\ref{sec:ablation_probe_noise} how much this
actually adds, and the answer is little: the five views prove to be near-duplicates, and the concatenation gains roughly 0.002 AUROC over the best single timestep. Since nothing is reconstructed, the decoder plays no part in any downstream task, and the extraction stops at the encoder levels without evaluating the bottleneck.}

\subsubsection{Dataset and Class Imbalance}

We utilized the complete TUHSZ corpus for seizure detection, comprising three official splits: training (790,501 segments), development (550,597 segments), and evaluation (228,667 segments). The training and development sets were combined to maximize available data for fine-tuning, yielding 1,341,098 five-second EEG segments. The predefined evaluation set of the data set was held out for the final patient-wise assessment.

The dataset exhibits substantial class imbalance reflective of clinical reality:
\begin{itemize}
\item Training: 89,345 seizure vs. 701,156 non-seizure (11.3\% seizure)
\item Development: 28,945 seizure vs. 521,652 non-seizure (5.3\% seizure)
\item Evaluation: 15,361 seizure vs. 213,306 non-seizure (6.7\% seizure)
\end{itemize}

The combined training pool (train + dev) contained 118,290 seizure and 1,222,808 non-seizure segments (8.8\% seizure prevalence). A stratified 95\%/5\% split was applied to create training and validation partitions while preserving class distributions. To address the class imbalance, we employed inverse frequency weighting in the cross-entropy loss:
\begin{equation}
w_k = \frac{N}{2 \cdot n_k}
\end{equation}
where $N$ is the total sample count and $n_k$ is the count for class $k$.

\subsubsection{Reinforced Decision Layer}
\label{sec:rdl}

Standard cross-entropy training optimizes per-sample accuracy, which can lead to classifiers that achieve high overall accuracy by favoring the majority class while underperforming on the clinically critical seizure class. To address this, we introduce a RDL that directly optimizes for seizure detection F1-score through PL learning \cite{he2022reinforcement}.

\hlIII{\textit{Attribution.} The individual components of this layer are not new and we do
not claim them. Policy-gradient optimisation of a non-differentiable objective through
REINFORCE with a moving-average baseline is standard~\cite{williams1992simple}, and
reinforcement learning has been applied previously to imbalanced medical
classification~\cite{he2022reinforcement}. What is specific to this work is the
composition: an antisymmetric, input-conditioned logit adjustment placed on top of
multi-timestep diffusion features and rewarded with the same F1 metric used for
evaluation. We also report, in Section~\ref{sec:ablation_logitadj}, that the measurable
contribution of this composition under our protocol is equivalent to learning a decision
threshold rather than improving the ranking of segments, and we present it on that basis
rather than as a source of additional discriminative power.}

\paragraph{Architecture}
The decision layer parameterizes a policy network $\pi_\phi: \mathbb{R}^D \rightarrow \mathbb{R}$ that maps the aggregated feature vector to a scalar logit adjustment:
\begin{equation}
a = \pi_\phi(\mathbf{z}) = \mathbf{W}_2 \cdot \text{ReLU}(\mathbf{W}_1 \mathbf{z} + \mathbf{b}_1) + \mathbf{b}_2
\end{equation}
where the hidden dimension is set to $\max(64, D/4)$. This adjustment is applied asymmetrically to the classifier logits:
\begin{equation}
\tilde{l}_{\text{seizure}} = l_{\text{seizure}} + a, \quad \tilde{l}_{\text{normal}} = l_{\text{normal}} - a
\end{equation}
Positive adjustments $a > 0$ increase sensitivity toward seizure detection, while negative adjustments increase specificity. Crucially, the network learns to modulate this trade-off conditioned on the input features, enabling context-dependent decision boundaries.

\paragraph{Policy Gradient Training}
\hl{To enable end-to-end gradient-based optimization of the decision boundary through policy gradient learning, we parameterize the classification decision as a Bernoulli random variable.} During training, we sample binary decisions from a Bernoulli distribution parameterized by the adjusted probabilities:
\begin{equation}
p = \sigma(\tilde{l}_{\text{seizure}} - \tilde{l}_{\text{normal}}), \quad \hat{y} \sim \text{Bernoulli}(p)
\end{equation}

\hl{This stochastic sampling strategy, standard in policy-gradient reinforcement learning \cite{williams1992simple}, offers two key advantages: (1) it decouples the decision boundary (parameterized by $p$) from the loss function, enabling the policy gradient to directly optimize for the target metric (F1-score) rather than cross-entropy; (2) the inherent stochasticity encourages exploration of the decision boundary during training, helping avoid local optima in imbalanced settings where the optimal decision threshold may differ substantially from the default 0.5. During inference, we employ deterministic threshold $p > 0.5$ for computational efficiency and reproducibility.}
The reward signal is the batch-level F1-score for the seizure class, computed on the sampled predictions:
\begin{equation}
R = \frac{2 \cdot \text{TP}}{2 \cdot \text{TP} + \text{FP} + \text{FN} + \epsilon}
\end{equation}
where TP, FP, and FN are accumulated across the minibatch and $\epsilon = 10^{-6}$ ensures numerical stability.

We employ the REINFORCE algorithm~\cite{he2022reinforcement} with a moving-average baseline for variance reduction. The policy gradient loss is:
\begin{equation}
\mathcal{L}_{\text{PG}} = -(R - b) \cdot \frac{1}{N_b} \sum_{i=1}^{N_b} \log p(\hat{y}_i | \mathbf{z}_i)
\end{equation}
where the baseline $b$ tracks the expected reward via exponential moving average:
\begin{equation}
b \leftarrow \alpha \cdot b + (1 - \alpha) \cdot R, \quad \alpha = 0.9
\end{equation}
This baseline subtracts the expected reward from the observed reward, reducing gradient variance while preserving the unbiased gradient estimate. Actions that yield above-average F1-scores receive positive gradient updates, reinforcing the policy toward improved seizure detection.

\subsubsection{Combined Training Objective}
\label{sec:combined_objective}

The total loss combines weighted cross-entropy with the policy gradient term:
\begin{equation}
\mathcal{L} = \mathcal{L}_{\text{CE}} + \lambda \mathcal{L}_{\text{PG}}
\end{equation}
The cross-entropy loss provides stable gradient signal for learning discriminative features, while the policy gradient loss directly optimizes the target clinical metric. \hlIII{The
weight $\lambda$ and the reward differ between the two downstream tasks and are stated in
full in Table~\ref{tab:rl_config}: binary detection uses $\lambda = 0.10$ with the
seizure-class F1 as reward, whereas the four-class subtype task uses $\lambda = 0.15$
with the macro F1, a five-epoch warmup during which the policy-gradient term is disabled,
and an entropy bonus of $\beta = 0.01$ added to $\mathcal{L}_{\text{PG}}$ to maintain
exploration over the larger action space. In each case the reward is the metric that task
is evaluated on, which is the property the decision layer is intended to exploit.} Gradients from both terms are clipped to unit norm before the optimizer step to ensure training stability.

\subsubsection{Fine-Tuning Protocol}

We adopt partial fine-tuning to balance representation preservation with task adaptation. The diffusion backbone is partitioned as follows:
\begin{itemize}
\item \textbf{Frozen:} Initial convolution, early downsampling blocks (levels 1--2), and first bottleneck block
\item \textbf{Trainable:} Final two downsampling blocks, last two bottleneck blocks, classification head, and reinforced decision layer.
\end{itemize}

\hlIII{One point about this partition should be made explicit. The configuration lists
the last two bottleneck blocks among the trainable layers, but feature extraction stops at
the encoder and never evaluates the bottleneck, so those parameters receive no gradient and
stay at their pre-trained values. What is actually adapted is the last encoder level, the
downsampling convolution before it, the classification head and the decision layer:
about 1.6M backbone parameters rather than the 3.4M the configuration nominally selects.}

\subsection{Seizure Subtypes Classification with Few-Shot Learning (K-Shot Analysis, without RL)}
To evaluate the transferability of the representations learned in stages 1 and 2, we conduct a few-shot learning (FSL) analysis on seizure subtypes. A key technical contribution of our framework is the reuse of the discriminative head from the binary detection task. Rather than initializing a stochastic classification head, we utilize the specific weights from the stage 2 detector (Section~\ref{sec:detection}). 
We found that the binary detector had already learned to distinguish ictal morphologies from the background manifold; therefore, by removing only the final binary output layer and replacing it with a $5$-class linear layer, the model preserved high-level discriminative features that a randomly initialized head would require significantly more data to learn. This architectural reuse demonstrated significant performance gains over a standard initialization approach.

EEG segments were first normalized using precomputed channel-wise mean and standard deviation. All available subtype folders were loaded, and the three least frequent subtype classes were excluded to reduce the impact of extreme class sparsity, leaving the five most represented classes for downstream classification as shown in Table~\ref{tab:subtype_distribution}. The retained subtype labels were remapped to contiguous class indices.

\begin{table}[h]
\centering
\caption{Seizure Subtype Class Distribution. \hlIII{All five classes are used in the segment-wise task (Tables~\ref{tab:full_supervised_subtype} and~\ref{tab:segment_wise_results}, Fig.~\ref{fig:k_shots_results}); the patient-wise task uses the first four and excludes ABSZ (Section~\ref{sec:class_selection}).}}
\label{tab:subtype_distribution}
\begin{tabular}{@{}lcc@{}}
\toprule
\textbf{Class} & \textbf{Subtype} & \textbf{Samples} \\
\midrule
Class 0 & FNSZ (Focal Non-Specific) & 68,369 \\
Class 1 & GNSZ (Generalized Non-Specific) & 43,399 \\
Class 2 & TNSZ (Tonic) & 2,276 \\
Class 3 & CPSZ (Complex Partial) & 16,761 \\
Class 4 & ABSZ (Absence) & 856 \\
\midrule
\textbf{Total} & \textbf{5 Classes} & \textbf{131,661} \\
\bottomrule
\end{tabular}
\end{table}

The dataset was split using a stratified 80/20 partition to preserve class proportions. To simulate few-shot learning, we sampled a fixed number of training examples per class, denoted by 
$K$ and evaluated multiple values of 
$K$. The tested values were constrained so that $K$ did not exceed half of the smallest training-class size, thereby avoiding overuse of the rarest class. For each training configuration, the model therefore used approximately $5K$ base training samples, with additional augmented samples generated only for minority classes. Data augmentation consisted of additive Gaussian perturbation applied to sampled training examples from classes with fewer than 2000 available samples.

The classifier architecture used a pretrained diffusion model's encoder as the feature extractor. Feature representations were obtained by probing the encoder at five diffusion timesteps 
\begin{equation}
\mathcal{T}_{probe} = \{50, 250, 500, 750, 950\}
\end{equation} At each timestep, multi-scale encoder activations were globally average-pooled and concatenated to form a fixed-dimensional representation. This representation was passed to a three-layer fully connected classifier with batch normalization, GELU activation, and dropout.

Two transfer-learning settings were compared: (i) a frozen-backbone setting, in which only the classifier head was optimized, and (ii) a partially unfrozen setting, in which the final two encoder blocks and the temporal embedding module were fine-tuned. In addition to the diffusion-pretrained backbone, the classifier head was optionally initialized from a previously trained seizure-versus-nonseizure model, excluding the final output layer, which was replaced to match the multi-class subtype classification task.

\subsection{Patient-Wise Seizure Subtypes Classification (with RL)}

Building upon the pre-trained diffusion model, we developed a downstream classification framework for seizure subtype identification. This section details the architectural modifications, training methodology, and evaluation protocol.

\subsubsection{Problem Formulation and Class Selection}
\label{sec:class_selection}

\hlIII{Distinguishing seizure subtypes, rather than only detecting that a seizure has
occurred, carries direct clinical consequences. Whether an event is focal or generalised
constrains the differential diagnosis and localises the epileptogenic zone, which is a
prerequisite for pre-surgical evaluation; it also guides pharmacological choice, since
several anti-seizure medications are effective for one class and can aggravate the other.
Absence and tonic events differ again in prognosis and in the monitoring they warrant.
A detector that reports only seizure or non-seizure leaves these decisions to manual
review of the raw recording, which is the bottleneck that automation is intended to
relieve. Subtype classification is correspondingly harder than detection: the classes are
separated by finer morphological differences and are far more unevenly represented, which
is why we report it separately and under both segment-wise and patient-wise protocols.}

Seizure subtype classification was formulated as a multi-class supervised learning problem. From the five most prevalent seizure types in the TUHSZ corpus (Table~\ref{tab:subtype_distribution}), we selected four classes based on sufficient representation: focal non-specific seizures (FNSZ, $n=68{,}369$), generalized non-specific seizures (GNSZ, $n=43{,}399$), complex partial seizures (CPSZ, $n=16{,}761$), and tonic seizures (TNSZ, $n=2{,}276$). Absence seizures (ABSZ, $n=856$) were excluded due to limited samples, which would compromise reliable cross-validation estimates. The resulting dataset comprised 130{,}805 five-second EEG segments from 279 patients.

\hlIII{The manuscript therefore reports two subtype tasks, and the number of classes is fixed by the evaluation protocol. The \emph{segment-wise} task uses all five classes of Table~\ref{tab:subtype_distribution}, including ABSZ (131{,}661 segments; Tables~\ref{tab:full_supervised_subtype} and~\ref{tab:segment_wise_results}, Fig.~\ref{fig:k_shots_results}). The \emph{patient-wise} task uses four classes, FNSZ, GNSZ, CPSZ and TNSZ (130{,}805 segments; Tables~\ref{tab:cv_results} and~\ref{tab:patient_wise_results_updated}). ABSZ cannot be carried into the patient-wise protocol: all 856 ABSZ segments come from only two of the 279 patients, so under a patient-disjoint five-fold partition the class is absent from the held-out patients of three folds and its per-class performance could not be estimated in those folds. A segment-level split does not have this constraint, which is why ABSZ is retained there.}

\subsubsection{Feature Extraction via Multi-Timestep Diffusion Probing}

Diffusion-based representation learning motivates probing a denoiser at several points along its noise schedule rather than at one~\cite{nichol2021improved}, and we follow that construction: features are extracted at multiple probe timesteps and concatenated. \hlIII{In this setting, the segment is passed to the encoder unmodified at every probe timestep; the index conditions the network through the FiLM pathway but no corruption is applied, so the timesteps do not correspond to different noise levels of the input. Section~\ref{sec:ablation_probe_noise} shows this is the correct choice, since corrupting the input as the generative process does costs 0.0654 AUROC and falls to chance at the highest timesteps. Second, the resulting five views are highly redundant rather than complementary: they differ by at most 0.0011 in probe AUROC and correlate above 0.995 pairwise, so the concatenation acts as a mild ensemble over near-duplicate views and improves on the best single timestep by roughly 0.002 AUROC. We report the five-timestep descriptor because it is what our results were produced with, and quantify both points in Section~\ref{sec:ablation_probe_noise}.}

Specifically, we define a set of probe timesteps $\mathcal{T} = \{50, 250, 500, 750, 950\}$ uniformly spanning the diffusion schedule ($T=1000$). For each input segment $\mathbf{x} \in \mathbb{R}^{C \times L}$ where $C=22$ channels and $L=1280$ samples (5 seconds at 256 Hz), we compute the encoder's hierarchical representations at each probe timestep. \hlIII{As above, the segment is passed to the encoder unmodified: the timestep index conditions the network but no noise is added to $\mathbf{x}$, and the forward pass is deterministic.} The diffusion backbone employs a U-Net architecture with channel multipliers $[1, 2, 4, 8]$ applied to a base channel dimension of 32, yielding feature maps at four resolution levels with dimensions $\{32, 64, 128, 256\}$ channels respectively.

At each timestep $t \in \mathcal{T}$, we extract the output of each encoder level $\ell \in \{1, 2, 3, 4\}$ after the residual blocks and apply global average pooling (GAP) to obtain a fixed-length representation:
\begin{equation}
    \mathbf{f}_\ell^{(t)} = \frac{1}{L_\ell} \sum_{i=1}^{L_\ell} \mathbf{h}_{\ell,i}^{(t)} \in \mathbb{R}^{d_\ell}
\end{equation}
where $\mathbf{h}_{\ell,i}^{(t)}$ denotes the hidden state at spatial position $i$ of level $\ell$ for timestep $t$, $L_\ell$ is the sequence length at level $\ell$ after downsampling, and $d_\ell \in \{32, 64, 128, 256\}$ is the channel dimension. The complete feature vector is formed by concatenating across all levels and timesteps:
\begin{equation}
    \mathbf{z} = \bigoplus_{t \in \mathcal{T}} \bigoplus_{\ell=1}^{4} \mathbf{f}_\ell^{(t)} \in \mathbb{R}^{D}
\end{equation}
where $D = |\mathcal{T}| \times \sum_{\ell} d_\ell = 5 \times (32 + 64 + 128 + 256) = 2{,}400$.

\begin{comment}
\subsubsection{Classification Head Architecture}

The extracted features are processed by a three-layer fully-connected classification head. The first layer projects the concatenated features to a hidden dimension of 512 units, followed by batch normalization, GELU activation, and dropout with probability $p_1 = 0.3$. The second layer reduces dimensionality to 256 units with identical normalization, activation, and dropout ($p_2 = 0.2$). The final layer produces logits for the four seizure subtypes:
%
% \begin{equation}
%     \hat{\mathbf{y}} = \mathbf{W}_3 \cdot \text{Dropout}_{p_2}\left(\text{GELU}\left(\text{BN}\left(\mathbf{W}_2 \cdot \text{Dropout}_{p_1}\left(\text{GELU}\left(\text{BN}\left(\mathbf{W}_1 \mathbf{z} + \mathbf{b}_1\right)\right)\right) + \mathbf{b}_2\right)\right)\right) + \mathbf{b}_3
% \end{equation}
%
\begin{equation}
\begin{aligned}
\hat{\mathbf{y}} &= \mathbf{W}_3 \cdot 
\text{Dropout}_{p_2} \Bigg( \\
&\quad \text{GELU} \Big( \\
&\quad\quad \text{BN} \Big( 
\mathbf{W}_2 \cdot 
\text{Dropout}_{p_1} \Big( \\
&\quad\quad\quad \text{GELU} \Big(
\text{BN} \big( \mathbf{W}_1 \mathbf{z} + \mathbf{b}_1 \big)
\Big) \Big) \\
&\quad\quad + \mathbf{b}_2
\Big) \Big) \\
&\quad \Bigg) + \mathbf{b}_3
\end{aligned}
\end{equation}
%
Class probabilities are obtained via softmax: $p(y=k|\mathbf{x}) = \text{softmax}(\hat{\mathbf{y}})_k$.
\end{comment}

\subsubsection{Hybrid Training Objective}

To address the dual challenges of class imbalance and optimizing for the target evaluation metric (F1-score), we employ a hybrid loss function combining weighted cross-entropy with a policy gradient auxiliary term (as in Sections~\ref{sec:rdl} and~\ref{sec:combined_objective}).

\subsubsection{Partial Fine-Tuning Strategy}
Transfer learning from the pre-trained diffusion model requires balancing preservation of learned EEG representations against task-specific adaptation. We adopt a partial fine-tuning strategy with differential learning rates.
\hlIII{The backbone is divided into three groups. The early layers, namely the input convolution and the first encoder levels that capture low-level temporal and spectral detail, stay frozen, so that the general signal features learned during pre-training are
preserved. The late layers, meaning the last encoder level and the downsampling
convolution that precedes it, are unfrozen and adapt to the task. The classification head
and decision layer are trained from scratch. In total, roughly 1.6M of the backbone's
9.7M parameters are updated, alongside the 1.4M-parameter head.}

We employ the AdamW optimizer~\cite{loshchilov2017decoupled} with differential learning rates: $\eta_{\text{head}} = 5 \times 10^{-4}$ for the classification head and $\eta_{\text{backbone}} = 1 \times 10^{-5}$ for unfrozen backbone layers. Weight decay of $10^{-4}$ is applied uniformly. The learning rate follows a cosine annealing schedule over the training horizon.

\subsubsection{Leave-one-Fold-Out: LOFO}
To obtain unbiased performance estimates and assess generalization across patients, we employed a 5-fold patient-wise cross-validation strategy, referred to as LOFO (Leave-One-Fold-Out). The 279 patients were randomly partitioned into five disjoint folds of approximately equal size (55--56 patients each). For each fold, one subset served as the validation set while the remaining four comprised the training set, ensuring strict patient-level separation to prevent data leakage. Figure 3 shows the patient-wise 5-fold cross-validation for seizure subtype
classification.

Models were fine-tuned for a maximum of 60 epochs with early stopping triggered after 20 consecutive epochs without improvement in validation weighted F1-score.
\begin{figure}[t]
\centering
\includegraphics[width=\linewidth]{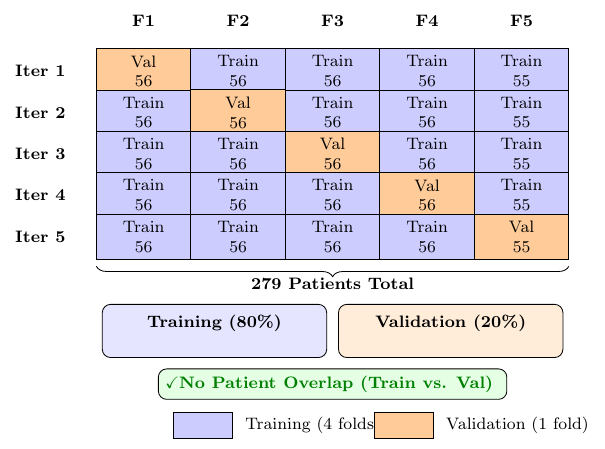}
\caption{Patient-wise 5-fold cross-validation for seizure subtype classification.}
\label{fig:patient_wise_cv}
\end{figure}

\subsection{Implementation Details}
\subsubsection{Data processing}
The proposed framework was developed and validated using the Temple University Seizure Corpus (TUHSZ v2.0.2), a large scale, open source collection of clinical EEG recordings. This study uses a publicly available, deidentified dataset, and no data sharing or IRB agreements are required to access the data. The dataset comprises anonymized recordings from over 600 patients, featuring diverse seizure morphologies across eight distinct seizure types. For the purposes of this study, we utilized the entire non seizure corpus for self supervised foundation pre training and reserved the annotated seizure segments for fine tuning and benchmarking.

To prepare the signals for deep learning, the raw EEG was segmented into 5-second windows with a 30\% overlap between consecutive segments to maintain temporal continuity. \hl{\textit{Segment Length Selection:} The 5-second duration (1,280 samples at 256 Hz) was selected based on two considerations: (1) \textit{Clinical relevance}: seizures typically exhibit discriminative EEG morphological changes within 2--5 second windows, allowing clinicians to make diagnostic decisions at sub-minute granularity; (2) \textit{Computational efficiency}: 5-second segments provide reasonable trade-off between temporal context required for seizure detection and computational tractability during training. This choice is tested empirically in Section~\ref{sec:window_size}.} Signal quality was ensured through a rigorous preprocessing pipeline consisting of a 50 Hz notch filter to suppress power-line interference, followed by a zero-phase Butterworth bandpass filter spanning 1 Hz to 75 Hz to remove low-frequency baseline drift and high-frequency muscular artifacts. Channel-wise normalization was finally applied using the global mean and standard deviation computed from the non-seizure pretraining set, ensuring numerical stability throughout the diffusion process.

\subsubsection{Training Environment and Foundation Model Implementation}
The DiffEEG framework was implemented using the PyTorch deep learning library. Due to the high computational overhead of 1D U-Net diffusion models and the scale of the TUHSZ dataset, all experiments were conducted on the Digital Research Alliance of Canada (formerly Compute Canada) High-Performance Computing (HPC) infrastructure.

A single NVIDIA A100 (80GB) GPUs per node was used for pre-training the foundation model (Stage 1). The complete pretraining architecture comprises approximately 9.6 million parameters, including 23 residual blocks (4 encoder levels $\times$ 2 blocks + 3 bottleneck + 
4 decoder levels $\times$ 3 blocks) and 3 strided downsampling convolutions. \hlIII{For downstream classification tasks the decoder pathway is discarded. Feature
extraction evaluates the input convolution, the shared timestep MLP and the four encoder
levels, and stops there: it comprises 3{,}396{,}160 parameters, of which the timestep MLP
accounts for 1{,}050{,}112. The bottleneck blocks (2{,}763{,}264 parameters) and the
decoder (3{,}512{,}032) are not evaluated when features are extracted.} 
\section{Results}
In this section, we present the empirical evaluation of the proposed DiffEEG framework.

\begin{figure*}[!t]
    \centering
    \includegraphics[width=0.6\textwidth]{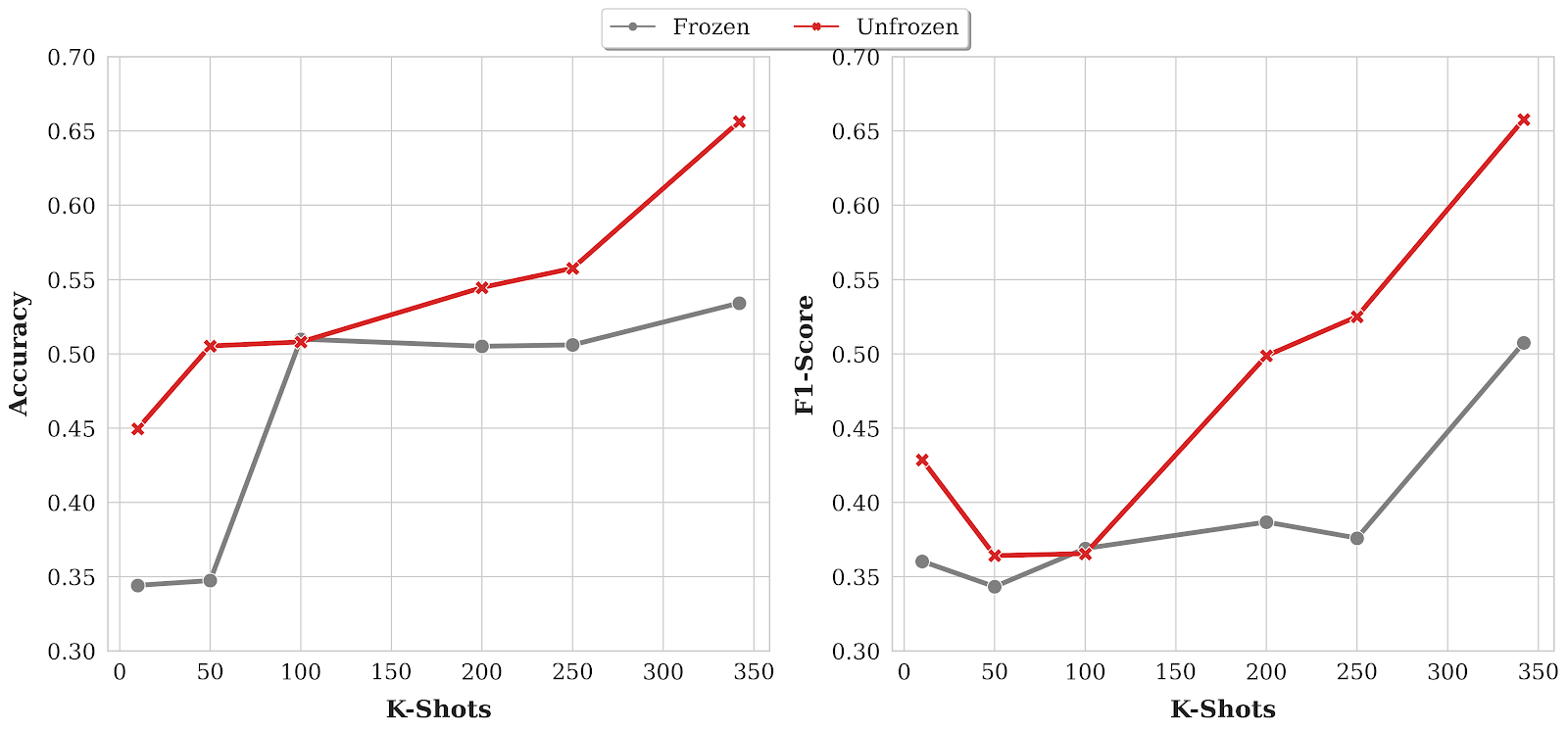}
    \caption{Few-Shot Seizure 5-Subtypes Classification Performance (Frozen vs. Unfrozen Backbone Strategies), 80/20 stratified split, segment-wise, without RL.}
    \label{fig:k_shots_results}
\end{figure*}

\hlIII{The evaluation spans four corpora and five downstream tasks, so that the
representation is not assessed on a single dataset. On \textbf{TUHSZ} we report binary
seizure detection under a patient-wise split (Table~\ref{tab:binary_detection}) and
seizure subtype classification under both a segment-wise protocol
(Table~\ref{tab:full_supervised_subtype}) and a patient-wise 5-fold protocol
(Table~\ref{tab:cv_results}), together with a few-shot analysis, and a controlled
comparison against three task-trained architectures that we train ourselves on the same
segments (Table~\ref{tab:adapted_tusz}). On \textbf{TUAB} we
report cross-dataset abnormality detection (Table~\ref{tab:tuab_bench}), a different
clinical question on a different corpus. On the \textbf{Bonn} database
(Table~\ref{tab:probe_bonn}) and the \textbf{Siena Scalp EEG} database
(Table~\ref{tab:probe_siena}) we report controlled external validation against five
foundation models, four of which appear on both corpora and one, EEGMamba, on Bonn
alone for the pre-training reason given in Section~\ref{sec:external_probe}. Both corpora
are external to DiffEEG: its pre-training uses only the non-seizure portion of TUHSZ, so
no seizure label from any corpus informs the representation. For the baselines we checked
each published pre-training corpus list rather than assuming, and two of them, LaBraM and
EEGMamba, name the Siena database among their pre-training data; the tables mark this
explicitly.}
\hlIII{The RL policy-gradient auxiliary loss of Section~\ref{sec:rdl} is used in binary seizure detection (Tables~\ref{tab:binary_detection} and~\ref{tab:adapted_tusz}), in patient-wise subtype classification (Tables~\ref{tab:cv_results} and~\ref{tab:patient_wise_results_updated}) and in TUAB fine-tuning. The segment-wise and few-shot subtype experiments (Tables~\ref{tab:full_supervised_subtype} and~\ref{tab:segment_wise_results}, Fig.~\ref{fig:k_shots_results}) and the linear-probe evaluations (Tables~\ref{tab:probe_bonn} and~\ref{tab:probe_siena}) are trained without it.} The RL configuration is shown in Table~\ref{tab:rl_config}.

\begin{table}[H]
\centering
\caption{\hlIII{Reinforced decision layer configuration. The two downstream tasks use
different settings; both are given here.
The reward is the metric each task is evaluated on, and the entropy bonus and warmup are
used only for the four-class task, where the policy acts over more than two outcomes.}}
\label{tab:rl_config}
\begin{tabular}{@{}lcc@{}}
\toprule
\textbf{Parameter} & \textbf{Binary detection} & \textbf{Subtype (LOFO)} \\
\midrule
RL weight ($\lambda$)         & 0.10 & 0.15 \\
Reward metric                 & Seizure-class F1 & Macro F1 \\
Baseline momentum ($\alpha$)  & 0.9  & 0.9 \\
Warmup epochs                 & --   & 5 \\
Entropy bonus ($\beta$)       & --   & 0.01 \\
Policy hidden width           & $\max(64, D/4)$ & $\max(64, D/4)$ \\
Baseline                      & Moving average & Moving average \\
\bottomrule
\end{tabular}
\end{table}
\subsection{Binary Seizure Detection Performance (with RL)}
The performance of the model fine-tuned for the binary seizure classification task was assessed on the held-out Evaluation set predefined in the TUHSZ dataset. As shown in Table~\ref{tab:binary_detection}, the model achieved a ROC-AUC of 0.7557 and a weighted F1-score of 0.85. The model achieved an overall accuracy of 0.81. While the specificity for the non-seizure class was high (Precision: 0.97, Recall: 0.82), the seizure class recall reached 0.59. 

\begin{table}[H]
\centering
\caption{Binary Seizure Detection Performance on Held-Out Evaluation Set (Patient-Wise Splits, with RL)}
\label{tab:binary_detection}
\begin{tabular}{lc}
\hline
\textbf{Metric} & \textbf{Evaluation Set (LOFO)} \\
\hline
Samples & 228,667 \\
Seizure Prevalence & 6.7\% \\
ROC-AUC & \textbf{0.7557} \\
Accuracy & 0.81 \\
Seizure Recall & 0.59 \\
Macro F1 & 0.59 \\
Weighted F1 & 0.85 \\
\hline
\end{tabular}

\end{table}
\subsection{Segment-Wise Seizure Subtype Classification (without RL)}
To evaluate the transferability of the learned representations, we conducted experiments on seizure subtype classification. We explored two distinct regimes: a few-shot learning regime (limited labels per class) and a full supervised sine-tuning regime (abundant labels).

For the few-shot subtype classification experiments, we selected the five most represented seizure subtype classes from the TUHSZ corpus to ensure sufficient sample support across all K-shot configurations. The resulting dataset comprised 131,661 five-second EEG segments distributed across five classes: Class 0 (FNSZ): 68,369 samples, Class 1 (GNSZ): 43,399 samples, Class 2 (TNSZ): 2,276 samples, Class 3 (CPSZ): 16,761 samples, and Class 4 (ABSZ): 856 samples. This selection strategy prioritized classes with adequate representation while maintaining clinical relevance across focal, generalized, and absence seizure categories.

\subsubsection{Few-Shot Learning (K-Shot Analysis)}
In this setting, we simulated low-label clinical scenarios by restricting the training data to samples per class. We compared frozen backbone strategies against unfrozen (partial fine-tuning) strategies.
As detailed in Figure 4, the unfrozen strategy consistently outperformed the frozen backbone. At the highest shot configuration (K=342), the unfrozen model achieved a Weighted F1-score of 0.6578. 

\subsubsection{Full Supervised Fine-Tuning (80/20 Stratified Split)}
\label{sec:supervised_split}
To establish an upper bound on performance when label scarcity is not a constraint, we fine-tuned the pretrained model on a larger subset of data using an 80/20 stratified train/test split. Unlike the binary detection task, this split was performed at the segment level, meaning segments from the same patient could appear in both training and testing sets. This setup evaluates the model's capacity to learn subtype distinctions when provided with sufficient data, albeit with potential patient-level data leakage.
The model demonstrated exceptional performance in this regime, achieving an Accuracy of 0.9766 and a Weighted F1-score of 0.9767. As shown in Table~\ref{tab:full_supervised_subtype}, the model maintained high precision and recall across all five major subtype classes, including the minority classes (e.g., Subtype 5, samples=856). 

\begin{table}[H]
\centering
\caption{Full Supervised Seizure 5-Subtypes Classification Performance (80/20 Stratified Split, segment-wise, without RL)}
\label{tab:full_supervised_subtype}
\small
\begin{tabular*}{\columnwidth}{@{}l@{\extracolsep{\fill}}ccc@{}}
\toprule
\textbf{Class} & \textbf{Prec} & \textbf{Rec} & \textbf{F1} \\
\midrule
Subtype 1 (FNSZ) & 0.9873 & 0.9702 & 0.9787  \\
Subtype 2 (GNSZ) & 0.9549 & 0.9827 & 0.9686 \\
Subtype 3 (CPSZ) & 0.9919 & 0.9848 & 0.9884 \\
Subtype 4 (TNSZ) & 0.9707 & 0.9886 & 0.9795 \\
Subtype 5 (ABSZ) & 0.9907 & 0.9953 & 0.9930 \\
\midrule
Accuracy         & --     & --     & 0.9766 \\
Macro Average    & 0.9791 & 0.9843 & 0.9816 \\
Weighted Average & 0.9770 & 0.9766 & 0.9767 \\
\bottomrule
\end{tabular*}
\end{table}

\subsection{Patient-Wise Seizure Subtype Classification (5-Fold Leave-One-Fold-Out Cross-Validation, LOFO)}
To rigorously evaluate the generalization capability of DiffEEG across unseen patients, we conducted a 5-fold patient-wise cross-validation experiment for seizure subtype classification, as shown in Figure 3. Unlike the stratified segment-level split described in Section~\ref{sec:supervised_split}, this protocol ensures strict patient-level separation: no EEG segments from validation patients appear during training, preventing data leakage that could arise from within-patient temporal correlations.

\subsubsection{LOFO Results}
Performance varied across folds, reflecting the inherent difficulty of cross-patient generalization in EEG analysis. Table~\ref{tab:cv_results} summarizes the best validation performance per fold.
The mean weighted F1-score across folds was 0.59, with an accuracy of 0.61. Fold 2 achieved the highest performance (F1: 0.6738), while Folds 1 and 5 showed lower performance (F1: 0.4866 and 0.4788, respectively). 
\begin{figure}[!t]
    \centering
    \includegraphics[width=\linewidth]{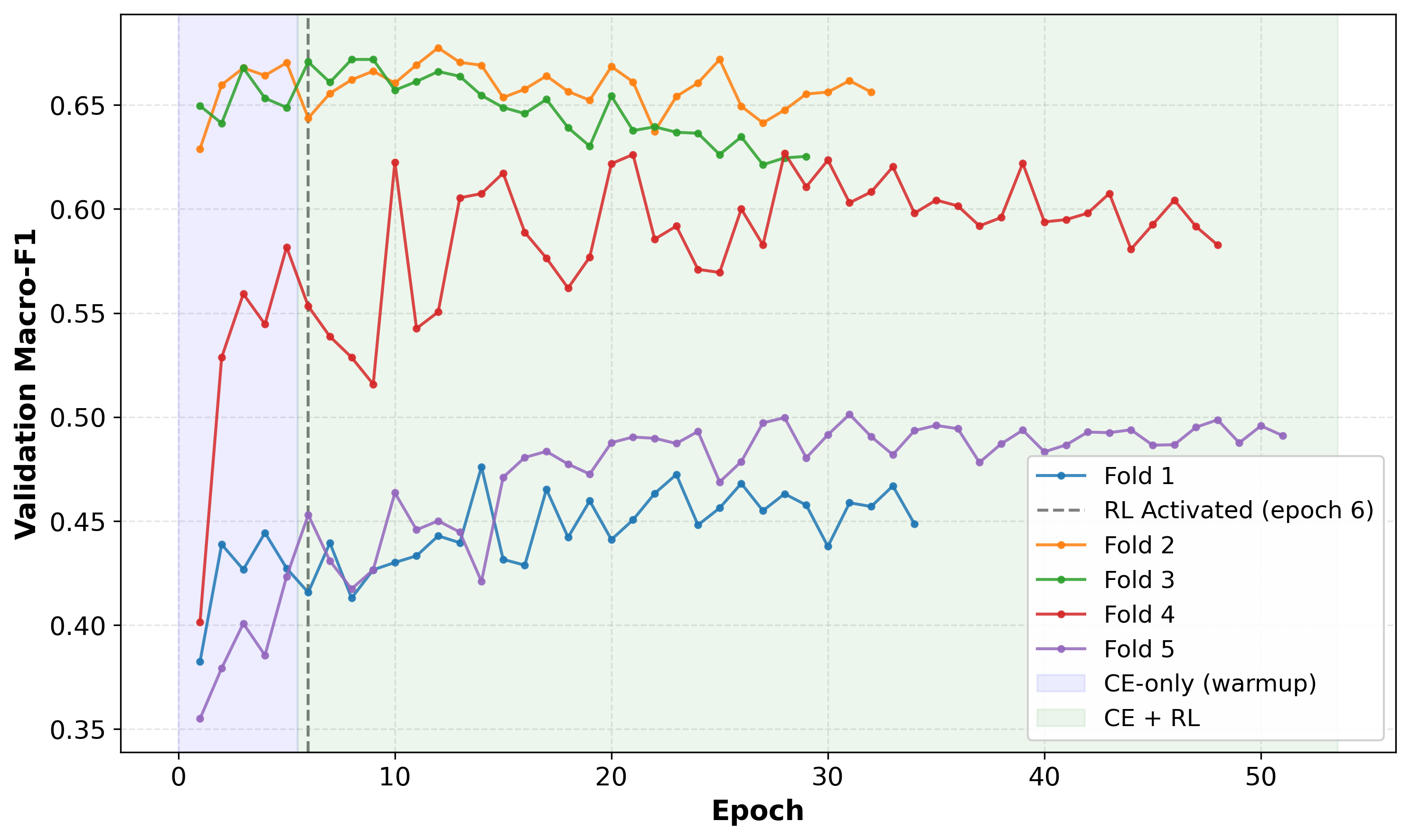}
    \caption{Fine-tuning and reinforcement learning dynamics for the proposed model. The vertical dashed line indicates the activation of the Policy Gradient-based decision layer at Epoch 6.}
    \label{fig:model_comparison}
\end{figure}
\begin{table}[t]
\centering
\caption{5-Fold Patient-Wise Cross-Validation Results (Seizure 4-Subtypes Classification)}
\label{tab:cv_results}
\footnotesize
\setlength{\tabcolsep}{3.5pt}
\begin{tabular}{@{}cccccc@{}}
\toprule
\textbf{Fold} & \textbf{Train} & \textbf{Test} & \textbf{Test Acc} & \textbf{Test F1} \\
\midrule
1 & 223 & 56 & 0.5156 & 0.4866 \\
2 & 223 & 56 & 0.6953 & 0.6738 \\
3 & 223 & 56 & 0.6661 & 0.6502 \\
4 & 223 & 56 & 0.6555 & 0.6580 \\
5 & 224 & 55 & 0.5023 & 0.4788 \\
\midrule
\textbf{Mean $\pm$ Std (with RL)} & \textbf{--} & \textbf{--}  &  \textbf{0.61 $\pm$ 0.0812} & \textbf{0.59 $\pm$ 0.0875} \\
\textbf{Mean $\pm$ Std (no RL)} & \textbf{--} & \textbf{--}  &  \textbf{0.62 $\pm$ 0.0748} & \textbf{0.60 $\pm$ 0.0827} \\
\bottomrule
\end{tabular}
\end{table}

\subsubsection{Training Dynamics and RL Contribution}
The policy gradient (RL) auxiliary loss was activated after epoch 5 (warmup period). As shown in Figure 5, the RL reward (macro F1) rose after activation, from 0.4617 at epoch 6 to 0.7461 at epoch 49 in Fold 2. \hlIII{The reward is the training macro F1, which rises as the model fits the training data whether or not the policy-gradient term is present, so the curve on its own does not establish that the term caused the improvement. The controlled ablations that do isolate it, in Sections~\ref{sec:ablation_logitadj} and~\ref{sec:window_size}, find its contribution to discrimination to be indistinguishable from zero. What the curve shows is that the objective is optimised as intended, not that optimising it improves the representation.}
Early stopping was triggered between epochs 23--49 across folds, with no improvement in validation F1 for 20 consecutive epochs. 
\subsection{Comparison and Benchmarking}
\label{sec:benchmarking}
To evaluate the architectural capacity of DiffEEG, we benchmarked the framework against state-of-the-art supervised baselines and large-scale EEG foundation models across three primary objectives. Table~\ref{tab:segment_wise_results} presents the comparative results of the segment-wise classification of the five seizure subtypes against other supervised deep learning models. Table~\ref{tab:patient_wise_results_updated} details the patient-wise seizure subtype classification performance in comparison to related CNN and transformer-based models. Furthermore, Table~\ref{tab:tuab_bench} reports a cross-dataset evaluation on the TUAB dataset for clinical abnormality detection against other foundational architectures.

\begin{table}[H]
\centering
\caption{Segment-Wise (Seizure-Wise) Seizure 5-Subtypes Classification Performance on TUHSZ. \hlIII{Entries from the literature are the values reported in their own publications and use the class groupings noted in the second column; our row is the 5-class result of Table~\ref{tab:full_supervised_subtype}.}}
\label{tab:segment_wise_results}
\resizebox{\columnwidth}{!}{ % Shrink to fit a single column
\begin{tabular}{llcc}
\toprule
\textbf{Study} & \textbf{Methodology} & \textbf{Acc. (\%)} & \textbf{W-F1 (\%)} \\
\midrule
Albaqami et al. (2022) \cite{Albaqami2022wavelet} & LightGBM + DTCWT (7-cl) & -- & 96.04 \\
ResBiLSTM (2024) \cite{Zhao2024residual} & ResNet + BiLSTM & 95.03 & 95.03 \\
Huang et al. (2023) \cite{huang2023multiband} & 3D-CBAMNet & 94.47 & 94.38 \\
Asif et al. (2020) \cite{asif2020seizurenet} & SeizureNet (Ens. CNNs) & -- & 94.00 \\
Jia et al. (2022) \cite{jia2022variable} & VWCNNs & 91.71 & 94.00 \\
Li et al. (2020) \cite{Li2020epileptic} & CE-stSENet & 92.00 & 93.69 \\
Zhang et al. (2022) \cite{zhang2022combination} & VMD + NLTWSVM & 92.29 & 92.30 \\
Roy et al. (2020) \cite{roy2020seizure} & k-NN / XGBoost & -- & 90.10 \\
\midrule
\textbf{DiffEEG} & \textbf{Denoising Diffusion (no RL)} & \textbf{97.66} & \textbf{97.67} \\
\bottomrule
\end{tabular}
}
\end{table}
\FloatBarrier
\renewcommand{\arraystretch}{0.7}
\begin{table*}[t]
\centering
\caption{Patient-Wise (LOFO) Seizure 4-Subtypes Classification Comparison on TUHSZ. \hlIII{Literature entries use the class groupings in the \emph{Classes} column.}}
\label{tab:patient_wise_results_updated}
\begin{tabular}{lccccc}
\toprule
\textbf{Study} & \textbf{TUHSZ Ver.} & \textbf{Classes} & \textbf{Patients} & \textbf{Accuracy (\%)} & \textbf{Weighted F1 (\%)} \\
\midrule
MBMD Transformer (2024) \cite{Peng2024multi} & v1.5.2 & 4 & 68 & 74.6 & 73.9 \\
Albaqami et al. (2022) \cite{Albaqami2022wavelet} & v1.5.2 & 5 & $>$300 & -- & 74.7 \\
Wavelet2Vec (2023) \cite{peng2023wavelet2vec} & v1.5.2 & 4 & 68 & 73.0 & 72.0 \\
TIE-EEGNet (2022) \cite{peng2022tie} & v1.5.2 & 4 & 68 & 63.0 & 66.0 \\
SeizureNet (2020) \cite{asif2020seizurenet} & v1.5.2 & 7 & ~315 & -- & 59.2 \\
Albaqami et al. (2022) \cite{Albaqami2022wavelet} & v1.5.2 & 7 & $>$300 & -- & 56.2 \\
Roy et al. (2020) \cite{roy2020seizure} & v1.5.2 & 7 & ~315 & -- & 56.1 \\
\midrule
\textbf{DiffEEG - Average (no RL)} & \textbf{v2.0.2} & \textbf{4} & \textbf{279} & \textbf{62.0} & \textbf{60.0} \\
\textbf{DiffEEG - Best Fold (Fold 2)} & \textbf{v2.0.2} & \textbf{4} & \textbf{56} & \textbf{70.01} & \textbf{67.78} \\
\bottomrule
\end{tabular}
\end{table*}

\subsubsection{\hlIII{Binary Seizure Detection Against Task-Trained Baselines (DiffEEG with RL)}}
\label{sec:adapted_tusz}
\hlIII{The benchmarking above compares DiffEEG against results reported by other authors.
To obtain a comparison in which we execute every model ourselves, we additionally trained
three widely used EEG architectures on the binary seizure-detection task and evaluated
them beside our own fine-tuned model on the same segments. The baselines are
EEGNet~\cite{lawhern2018eegnet}, EEG-Conformer~\cite{song2023conformer} and the
spatio-temporal transformer of Song et al.~\cite{song2022eeg}. None of the three
distributes pre-trained weights, so each is trained end-to-end from random
initialisation.}

\hlIII{\textbf{Protocol.} All four models receive identical inputs, splits, optimiser and
metrics; only the architecture and the adaptation differ. Inputs are the 22 monopolar
channels at 256\,Hz with channel-wise standardisation that DiffEEG consumes. Every model
is trained with AdamW, a three-epoch warmup followed by cosine decay, class-weighted
cross-entropy to counter the imbalance, and early stopping on validation AUROC with a
patience of eight epochs; the checkpoint with the best validation AUROC is evaluated once
on the held-out set. Confidence intervals are obtained from a 1{,}000-fold bootstrap of
that evaluation set. The three baselines train every parameter, whereas DiffEEG is
adapted by partial unfreezing of its pre-trained backbone together with the reinforced
decision layer, so the baselines receive at least as much task-specific adaptation as our
model does and the comparison is not weighted in our favour.}

\hlIII{\textbf{Data.} These experiments use a fixed subset of TUHSZ rather than the whole
corpus, drawn once with a fixed seed and identical for every model. The training set is
100{,}000 segments sampled class-balanced from the pooled train and development splits,
with a further 20{,}000 segments held out for validation, and the evaluation set is
50{,}000 segments drawn uniformly at random from the official \emph{eval} split so that
it retains the natural prevalence of about 6.7\,\% seizure (3{,}443 seizure segments in
the realised draw). Patient-level separation is
preserved, since the corpus splits are themselves patient-disjoint. Absolute values are
therefore not directly comparable with Table~\ref{tab:binary_detection}, which is computed
on all 228{,}667 evaluation segments; the comparison between rows within
Table~\ref{tab:adapted_tusz} is the quantity of interest.}

\begin{table}[htbp]
\centering
\caption{\hlIII{Binary seizure detection on TUHSZ, all models executed by us on an
identical 50{,}000-segment evaluation subset at the natural prevalence of about 6.7\,\%
seizure. The three baselines are trained end-to-end from random initialisation without RL; DiffEEG
is fine-tuned from its pre-trained backbone with RL (reinforced decision layer). Best value per column in bold. EEGNet's
accuracy and specificity should be read together with its sensitivity, discussed in the
text.}}
\label{tab:adapted_tusz}
\resizebox{\columnwidth}{!}{
\begin{tabular}{llccccccc}
\toprule
\textbf{Model} & \textbf{Adaptation} & \textbf{Params} & \textbf{Acc.} & \textbf{AUROC} & \textbf{PR-AUC} & \textbf{F1$_{\mathrm{m}}$} & \textbf{Sens.} & \textbf{Spec.} \\
 & & \textbf{(M)} &  &  &  &  &  &  \\
\midrule
EEGNet \cite{lawhern2018eegnet} & From scratch & 0.003 & \textbf{0.93} & \textbf{0.88} & 0.48 & 0.55 & 0.07 & \textbf{0.99} \\
EEG-Conformer \cite{song2022eeg} & From scratch & 0.97 & 0.77 & 0.77 & 0.26 & 0.57 & 0.65 & 0.78 \\
ST-Transformer \cite{song2021transformer} & From scratch & 3.43 & 0.78 & 0.83 & 0.40 & 0.59 & 0.74 & 0.78 \\
\midrule
\textbf{DiffEEG (Ours)} & Fine-tuned + RL & 9.67 & 0.83 & 0.86 & \textbf{0.50} & \textbf{0.64} & \textbf{0.75} & 0.84 \\
\bottomrule
\end{tabular}}
\end{table}

\hlIII{DiffEEG attains the best macro F1 (0.64), PR-AUC (0.50) and sensitivity
(0.75) of the four. Against the spatio-temporal transformer its confidence intervals
are disjoint, and its advantage therefore statistically significant, on accuracy
($+0.047$), AUROC ($+0.027$), PR-AUC ($+0.087$), macro F1 ($+0.044$) and specificity
($+0.050$), with sensitivity statistically tied; against EEG-Conformer it is significantly
better on all six metrics, including PR-AUC ($+0.227$).}

\hlIII{EEGNet requires separate comment, because its headline scores are misleading in
this setting. Its accuracy of 0.9345 is the highest in the table, but a detector that
never fires attains 0.9311 on this evaluation set, so the margin over a trivial classifier
is 0.34 percentage points. The reason is visible in the remaining columns: at a threshold
of 0.5 it operates at a specificity of 0.99 and a sensitivity of 0.07, that is, it
recovers roughly seven per cent of ictal segments. Its AUROC of 0.88 is genuinely the
highest of the four, so the representation ranks segments well; what fails is the
placement of the operating point, and the same pattern recurs in
Section~\ref{sec:ablation_logitadj}. DiffEEG exceeds it by $+0.676$ in sensitivity and
$+0.089$ in macro F1, both with disjoint intervals. This is the clearest illustration in
our results of why accuracy alone is an inadequate summary at clinical prevalence.}

\hlIII{Because Table~\ref{tab:adapted_tusz} evaluates DiffEEG after fine-tuning while
Section~\ref{sec:external_probe} evaluates it frozen, the same evaluation subset also lets
us isolate what adaptation contributes to our own model.
Table~\ref{tab:diffeeg_ft_gain} reports both configurations on those identical 50{,}000
segments.}

\begin{table}[htbp]
\centering
\caption{\hlIII{Effect of fine-tuning DiffEEG, measured on the same 50{,}000-segment
TUHSZ evaluation subset. The frozen row trains only a linear classifier on the
pre-trained representation; the fine-tuned row is the configuration this paper proposes.}}
\label{tab:diffeeg_ft_gain}
\resizebox{\columnwidth}{!}{
\begin{tabular}{lcccccc}
\toprule
\textbf{Adaptation of DiffEEG} & \textbf{Acc.} & \textbf{AUROC} & \textbf{PR-AUC} & \textbf{F1$_{\mathrm{m}}$} & \textbf{Sens.} & \textbf{Spec.} \\
\midrule
Frozen backbone, linear probe & 0.76 & 0.81 & 0.30 & 0.58 & 0.73 & 0.76 \\
Fine-tuned (partial unfreezing + RDL) & \textbf{0.83} & \textbf{0.86} & \textbf{0.49} & \textbf{0.64} & \textbf{0.75} & \textbf{0.83} \\
\midrule
\textit{Absolute gain from fine-tuning} & $+0.0689$ & $+0.0460$ & $+0.1913$ & $+0.0636$ & $+0.0220$ & $+0.0723$ \\
\bottomrule
\end{tabular}}
\end{table}

\hlIII{Adaptation improves every metric, and by a wide margin on the
prevalence-sensitive ones: PR-AUC rises by 0.191 and macro F1 by 0.064, with accuracy,
AUROC and specificity gaining between 0.046 and 0.072. The diffusion representation is
therefore a useful initialisation rather than a sufficient one on its own, and the
fine-tuning stage, not the frozen encoder, is what produces the performance this paper
reports.}

\subsubsection{Abnormal EEG Detection on TUAB}
\label{sec:tuab}
Table~\ref{tab:tuab_bench} shows the benchmarking results on the TUAB dataset, in which DiffEEG attains the highest AUROC of the models listed, 0.922, ahead of larger foundation models such as LaBraM (0.902) and BioSerenity-E1 (0.890). \hlIII{Three qualifications are needed. First, every competing entry is the value its own authors report, obtained under their own preprocessing, splits and fine-tuning budget; none was re-run by us, so no statistical comparison is possible and the word ``significant'' does not apply to this table. Second, the advantage is confined to AUROC. On PR-AUC, DiffEEG's 0.860 is last among the five entries that report it, behind BioSerenity-E1 (0.910), LaBraM (0.897), BIOT (0.879) and EEGFormer (0.872); EEGPT do not report AUC-PR on this task. Its accuracy is likewise mid-table. The fair reading is that DiffEEG ranks abnormal recordings well while several other models place their operating point better. Third, TUAB is not a seizure task, and it is the benchmark on which our representation is least aligned with its pre-training corpus; the controlled seizure comparisons of Section~\ref{sec:external_probe}, where we execute every model ourselves, carry more weight than this table does.}
\begin{table}[htbp]
\centering
\caption{Benchmarking Foundation Models on TUAB. \hlIII{Baseline values are those reported by their respective authors and were not re-run by us, so the rows are not protocol-matched and no significance can be claimed. Baselines report \emph{balanced} accuracy in the last column, whereas our entry is plain accuracy; TUAB is close to balanced, but the two are not the same quantity. EEGPT report no AUC-PR for this task, hence the dash. Best value per column in bold.}}
\label{tab:tuab_bench}
\resizebox{\columnwidth}{!}{
\begin{tabular}{llccc}
\toprule
\textbf{Model} & \textbf{Type} & \textbf{AUROC} & \textbf{PR-AUC} & \textbf{Acc / B-Acc} \\
\midrule
\textbf{DiffEEG (Ours)} & \textbf{Denoising Diffusion} & \textbf{0.92} & 0.86 & 80.00\% \\
BioSerenity-E1 (2025) \cite{bettinardi2025bioserenity} & VQ-VAE  & 0.89 & \textbf{0.91} & \textbf{82.25\%} \\
LaBraM (2024) \cite{jiang2024labram} & Neural Tokenizer  & 0.90 & 0.90 & 81.40\% \\
BIOT (2023) \cite{yang2023biot} & Biosignal Transformer & 0.88 & 0.88 & 79.59\% \\
EEGPT (2024) \cite{wang2024eegpt} & Hierarchical Transformer & 0.87 & -- & 79.83\% \\
EEGFormer (2023) \cite{wan2023eegformer} & Transformer Tokenizer & 0.87 & 0.87 & -- \\
\bottomrule
\end{tabular}
}
\end{table}

\subsubsection{\hlIII{Controlled External Validation on Bonn and Siena (Linear Probing, without RL)}}
\label{sec:external_probe}
\hlIII{The comparisons reported above are assembled from the results each method
publishes, obtained under different preprocessing pipelines, splits and fine-tuning
budgets. They locate DiffEEG within the literature but cannot isolate the contribution of
the learned representation. To provide a comparison in which the data and the protocol
are held fixed, we re-evaluated five foundation models inside our own pipeline on two
external seizure corpora: the Bonn database~\cite{andrzejak2001indications} and the Siena
Scalp EEG database~\cite{detti2020eeg}. The baselines are EEGDM~\cite{puah2025eegdm}, the
other diffusion-based EEG representation model, BIOT~\cite{yang2023biot},
LaBraM~\cite{jiang2024labram} and CBraMod~\cite{wang2025cbramod}, joined on Bonn by the
state-space model EEGMamba~\cite{wang2025eegmamba}.}

\hlIII{\textbf{Protocol: linear probing, no fine-tuning.} Every backbone is frozen and
used exactly as released, and \emph{no model, including ours, is fine-tuned}. Only a
linear classifier is trained on top of the frozen representation, so what is compared is
the linear separability of each pre-trained representation rather than the capacity of
any downstream head or the budget spent adapting it. All models receive the same
segments, the same splits and the same classifier family, and each is given the input
view its own authors specify: DiffEEG 22 monopolar channels at 256\,Hz with channel-wise
standardisation, EEGDM, BIOT and CBraMod the TCP bipolar montage at 200\,Hz under their
published scalings, and LaBraM 21 monopolar channels at 200\,Hz. Each bipolar pair is a difference
of two of our monopolar channels and the 256\,Hz to 200\,Hz conversion is a rational
resampling, so no view supplies information the others are denied. Regularisation
strength is selected on a held-out validation split independently for each model over a
ten-point logarithmic grid, and all confidence intervals come from a 1{,}000-fold
bootstrap of the evaluation set. Parameter counts in
Tables~\ref{tab:probe_bonn} and~\ref{tab:probe_siena} are those reported by the
respective authors; for EEGDM this is the full published model, whereas our probe
instantiates its released backbone with a linear head in place of its Latent Fusion
Transformer. CBraMod's authors do not state a parameter count, so the value marked
$\dagger$ is measured directly from their released checkpoint. \emph{Pre-train corpora}
counts the number of distinct datasets each released checkpoint was pre-trained on, as
stated by its authors. It is a measure of the diversity of the pre-training data, not of
its volume: a single large corpus such as TUEG or TUHSZ can contain more recorded hours
than several small ones combined, so the column should not be read as a proxy for
pre-training scale.}

\begin{figure}[htbp]
    \centering
    \includegraphics[width=0.7\linewidth]{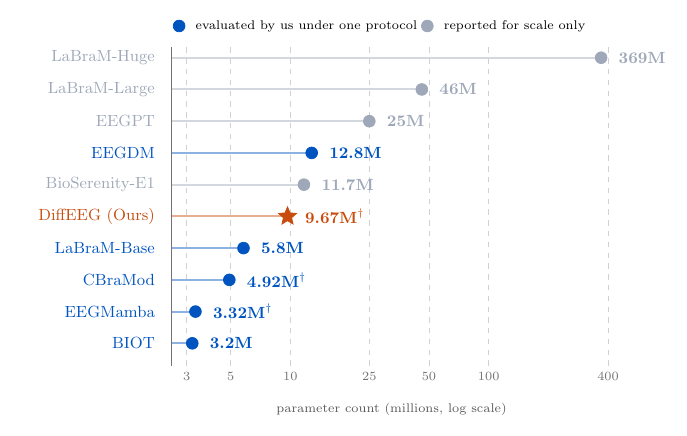}
    \caption{\hlIII{Parameter counts of DiffEEG and published EEG foundation models
    (log scale). Values are those reported by the respective authors, for the
    configurations they release.}}
    \label{fig:model_sizes}
\end{figure}

\hlIII{Figure~\ref{fig:model_sizes} places these parameter counts on the scale of
published EEG foundation models. DiffEEG sits in the middle of that range: an order of
magnitude smaller than the largest released configurations, and larger than the most
compact ones, including two of the baselines probed here.}

\begin{table}[htbp]
\centering
\caption{\hlIII{Frozen linear probe on the Bonn database (100 evaluation segments,
20\,\% seizure). All backbones are frozen; no model is fine-tuned and no RL is used. Best value per column
in bold, with ties shown.}}
\label{tab:probe_bonn}
\resizebox{\columnwidth}{!}{
\begin{tabular}{lcccccccccc}
\toprule
\textbf{Model} & \textbf{Params} & \textbf{Pre-train} & \textbf{Bonn} & \textbf{Acc.} & \textbf{AUROC} & \textbf{PR-AUC} & \textbf{F1$_{\mathrm{w}}$} & \textbf{F1$_{\mathrm{m}}$} & \textbf{Sens.} & \textbf{Spec.} \\
 & \textbf{(M)} & \textbf{corpora} & \textbf{seen?} &  &  &  &  &  &  &  \\
\midrule
EEGDM \cite{puah2025eegdm} & 12.8 & 1 & no & \textbf{0.94} & \textbf{0.99} & 0.97 & \textbf{0.94} & \textbf{0.91} & \textbf{1.00} & 0.92 \\
BIOT \cite{yang2023biot} & 3.2 & 6 & no & 0.93 & 0.97 & 0.94 & 0.93 & 0.89 & 0.95 & 0.92 \\
LaBraM \cite{jiang2024labram} & 5.8 & $\sim$20 & no & 0.92 & 0.99 & \textbf{0.98} & 0.92 & 0.89 & \textbf{1.00} & 0.90 \\
CBraMod \cite{wang2025cbramod} & 4.92$^\dagger$ & 1 & no & 0.93 & \textbf{0.99} & \textbf{0.98} & 0.93 & 0.90 & \textbf{1.00} & 0.91 \\
EEGMamba \cite{wang2025eegmamba} & 3.32 & 5 & no & \textbf{0.94} & 0.95 & 0.92 & \textbf{0.94} & \textbf{0.91} & 0.85 & \textbf{0.96} \\
\midrule
\textbf{DiffEEG (Ours)} & 9.67 & 1 & no & \textbf{0.94} & 0.98 & 0.94 & \textbf{0.94} & \textbf{0.91} & 0.95 & 0.94 \\
\bottomrule
\end{tabular}}
\end{table}

\hlIII{On Bonn all six representations are strong and, critically, mutually
indistinguishable: every 95\,\% confidence interval overlaps between every pair of models
on every metric, without exception. DiffEEG attains the highest accuracy (0.94, tied with
EEGDM and EEGMamba) together with the best weighted and macro F1, CBraMod attains the
highest AUROC and PR-AUC, and EEGMamba the highest specificity, but with 100 evaluation
segments the intervals span roughly $\pm$0.045 and no ordering can be asserted. Against
EEGMamba specifically, DiffEEG leads on five of the seven metrics, most visibly AUROC
($+0.034$) and sensitivity ($+0.100$), ties on accuracy and trails on specificity
($-0.025$); not one of those differences is significant, so the comparison establishes
parity rather than an advantage. Two further limitations should be stated plainly. First, the
Bonn recordings are single-channel, so in our common 22-channel layout only one channel
carries signal; monopolar models therefore see one live channel while the bipolar
montage yields two derived channels that are duplicates of it, and the corpus cannot
separate representation quality from that asymmetry. Second, with 360 training segments
the task is close to saturated for all six models. Bonn is therefore reported as a
sanity check confirming that DiffEEG transfers competitively to an unseen seizure
corpus, not as evidence of superiority.}

\begin{table}[htbp]
\centering
\caption{\hlIII{Frozen linear probe on the Siena Scalp EEG database (18{,}729 evaluation
segments at the corpus prevalence of 0.705\,\% seizure). All backbones are frozen; no
model is fine-tuned and no RL is used. \emph{Siena seen?} records whether this evaluation corpus appears in
that model's published pre-training data. Two models are exposed: the Siena Scalp EEG
Database is listed in LaBraM's pre-training appendix (30.47 hours) and in EEGMamba's
pre-training configuration, so for them Siena is not an external corpus. They are grouped
separately for that reason, and bold marks the best value per column \emph{among the
models for which Siena is external}.}}
\label{tab:probe_siena}
\resizebox{\columnwidth}{!}{
\begin{tabular}{lcccccccccc}
\toprule
\textbf{Model} & \textbf{Params} & \textbf{Pre-train} & \textbf{Siena} & \textbf{Acc.} & \textbf{AUROC} & \textbf{PR-AUC} & \textbf{F1$_{\mathrm{w}}$} & \textbf{F1$_{\mathrm{m}}$} & \textbf{Sens.} & \textbf{Spec.} \\
 & \textbf{(M)} & \textbf{corpora} & \textbf{seen?} &  &  &  &  &  &  &  \\
\midrule
EEGDM \cite{puah2025eegdm} & 12.8 & 1 & no & 0.91 & \textbf{0.91} & \textbf{0.24} & \textbf{0.95} & \textbf{0.53} & \textbf{0.74} & 0.91 \\
BIOT \cite{yang2023biot} & 3.2 & 6 & no & 0.78 & 0.70 & 0.02 & 0.87 & 0.45 & 0.54 & 0.78 \\
CBraMod \cite{wang2025cbramod} & 4.92$^\dagger$ & 1 & no & 0.87 & 0.85 & 0.18 & 0.92 & 0.50 & 0.70 & 0.87 \\
\midrule
LaBraM \cite{jiang2024labram} & 5.8 & $\sim$20 & \emph{yes} & 0.83 & 0.79 & 0.05 & 0.90 & 0.47 & 0.58 & 0.83 \\
EEGMamba \cite{wang2025eegmamba} & 3.32 & 5 & \emph{yes} & 0.83 & 0.85 & 0.25 & 0.90 & 0.48 & 0.77 & 0.83 \\
\midrule
\textbf{DiffEEG (Ours)} & 9.67 & 1 & no & \textbf{0.92} & 0.78 & 0.06 & \textbf{0.95} & \textbf{0.53} & 0.52 & \textbf{0.92} \\
\bottomrule
\end{tabular}}
\end{table}

\hlIII{Siena is the more demanding and more informative of the two: 18{,}729 evaluation
segments retained at the corpus prevalence of 0.705\,\% seizure, which is the regime in
which continuous monitoring actually operates. Here DiffEEG attains the best accuracy,
weighted F1 and specificity of every model compared, but ranks low on AUROC and PR-AUC,
and the comparison separates into distinct patterns.}

\hlIII{Two of the six models were pre-trained on Siena itself, which the Table~\ref{tab:probe_siena} marks. That exposure is worth reading closely,
because it bounds what pre-training on a target corpus actually buys. EEGMamba, which has
seen Siena, attains the highest PR-AUC (0.25) and sensitivity (0.77) in the table, the
shape one would predict; yet its PR-AUC exceeds EEGDM's by 0.003, and EEGDM never saw the
corpus, so the margin is far from significant. LaBraM is the more striking case: it was
also pre-trained on Siena and still falls significantly behind DiffEEG on accuracy
($+0.087$ in our favour), weighted F1 ($+0.050$), macro F1 ($+0.044$) and specificity
($+0.088$). Prior exposure to the evaluation corpus therefore did not secure an advantage
over a model pre-trained on a single unrelated corpus, which strengthens rather than
weakens the transfer claim we make for DiffEEG.}

\hlIII{Against BIOT and LaBraM the advantage is one-sided. Its confidence intervals are
disjoint from BIOT's, and its advantage therefore statistically significant, on accuracy
($+0.141$), PR-AUC ($+0.037$), weighted F1 ($+0.082$), macro F1 ($+0.067$) and
specificity ($+0.142$), and it is not significantly worse on any metric; against LaBraM it
is significantly better on accuracy ($+0.087$), weighted F1 ($+0.050$), macro F1
($+0.044$) and specificity ($+0.088$), again without being significantly worse anywhere.}

\hlIII{Against EEGDM and CBraMod the picture is mixed, and in the same direction in both
cases. DiffEEG is statistically tied with EEGDM on accuracy, weighted F1, macro F1 and
specificity while EEGDM is significantly stronger on AUROC ($+0.138$), PR-AUC ($+0.191$)
and sensitivity ($+0.220$). Against CBraMod, DiffEEG is significantly better on accuracy
($+0.048$), weighted F1 ($+0.027$), macro F1 ($+0.020$) and specificity ($+0.050$), while
CBraMod is significantly better on PR-AUC ($+0.132$) and sensitivity ($+0.182$), with
AUROC statistically tied. The consistent reading across both is that DiffEEG places its
operating point at least as well as any model compared here, but that two of the four
baselines order segments better, and CBraMod does so with a 200-dimensional descriptor
from a 4.92M-parameter backbone against DiffEEG's 2{,}400 dimensions from 9.67M.}

\hlIII{Two cautions govern how Table~\ref{tab:probe_siena} should be read. At a
prevalence of 0.705\,\% accuracy and specificity are weak evidence, since a detector that
never fires attains 99.3\,\% accuracy; the informative quantities are the
prevalence-sensitive ones, where chance-level PR-AUC is 0.007 and the models achieve
0.06 (DiffEEG), 0.24 (EEGDM), 0.18 (CBraMod), 0.05 (LaBraM) and 0.02 (BIOT), that is
between roughly 3 and 35 times chance. On PR-AUC specifically DiffEEG significantly
exceeds BIOT, is statistically tied with LaBraM, and is significantly exceeded by both
EEGDM and CBraMod. Equally, macro F1 lies between 0.45 and 0.53 for all five
models, so none of them solves this corpus; the appropriate claim is that DiffEEG is
competitive on a hard, realistically imbalanced rare-event task, not that it performs
well in absolute terms.}

\section{Discussion}

\hl{\subsection{Architectural and Design Justifications}

The architectural choices underlying DiffEEG are summarized below, together with the reasoning that determines each of them:

\subsubsection{Progressive Channel Expansion and Hierarchical Learning}
The encoder's channel multipliers [1, 2, 4, 8] are not arbitrary but enable hierarchical multi-scale representation learning. By expanding channels as temporal resolution decreases (32 $\rightarrow$ 64 $\rightarrow$ 128 $\rightarrow$ 256), the network captures increasingly abstract features while exponentially growing receptive fields. This is essential for EEG where discriminative information exists across multiple temporal frequencies (spike morphology at milliseconds, rhythm changes at seconds, state evolution at minutes). Downsampling, rather than being avoidable, is critical for expanding receptive fields and capturing long-range temporal dependencies spanning multiple seconds, exactly what seizure detection requires.

\subsubsection{Bottleneck Configuration (22 channels, 3 Blocks, $L/8$ Compression)}
\hlIII{As discussed in Section~\ref{sec:position}, the embedding width $D=512$ was aligned to the channel schedule rather than tuned: FiLM conditioning consumes $2C$ coefficients per block, the widest level carries $C=8m=256$ channels, and $D = 2\max_l C_l = 512$ makes the deepest conditioning projection square (Eq.~\eqref{eq:embed_dim}). We did not test smaller widths, and do not claim that one would perform worse.} \hl{The bottleneck operates on the $L/8$ temporal resolution produced by the three stride-2 downsampling stages implied by the four-level multiplier schedule $[1,2,4,8]$; three residual blocks at this resolution give the self-attention layers sufficient depth to aggregate global context.}

\subsubsection{Linear Interpolation plus Learned Convolution for Upsampling}
\hl{The decoder upsamples by linear interpolation followed by a learned $3\times1$ convolution. This is not restrictive for non-linear EEG signals: interpolation fixes only the resampling grid, while the subsequent learned convolution supplies the non-linear, signal-specific refinement and corrects interpolation artefacts.}

\hlIII{The reason for preferring this over a strided transposed convolution is worth
spelling out, because the term \emph{checkerboard artefact} is often used without its
mechanism. A transposed convolution with kernel size $k$ and stride $s$ scatters each
input value across $k$ consecutive output positions. When $k$ is not divisible by $s$,
different output positions end up receiving different numbers of contributions, and the
pattern repeats with period $s$. For the common setting $k=3$, $s=2$ the interior of the
output alternates between two contributions and one, so the operator imposes a fixed
$2\!:\!1$ amplitude modulation on its own output before any weight is learned. In a
two-dimensional image this alternation tiles along both axes and appears as a
checkerboard, which is where the name comes from~\cite{odena2016deconvolution}. In a
one-dimensional signal, it appears instead as a periodic
amplitude ripple at the stride period, a regular rise and fall every $s$ samples that the
network must then spend capacity suppressing. That is an unattractive failure mode for a
model whose output is itself an EEG waveform, where periodic amplitude structure is the
kind of input a downstream classifier may interpret as signal. Replacing the
operator with interpolation followed by a stride-one convolution removes the asymmetry at
the source: every output position is then reached by the same number of kernel taps.}

\hlIII{This resize-then-convolve construction is the standard remedy and is what the
diffusion literature adopts. DDPM upsamples by nearest-neighbour resizing followed by a
$3\times3$ stride-one convolution~\cite{ho2020denoising}, and the same pattern is used by
the ADM U-Net~\cite{dhariwal2021diffusion}, by the latent-diffusion
U-Net~\cite{rombach2022high}, and in the one-dimensional setting closest to ours by
WaveGrad~\cite{chen2021wavegrad}. We depart from those implementations in one respect
that should be stated: they interpolate with nearest-neighbour
resampling and we interpolate linearly. Both give the uniform overlap that avoids the
artefact, so the argument above is unaffected. Linear resampling additionally avoids the
piecewise-constant staircase that nearest-neighbour upsampling produces, whose stepwise
discontinuities carry energy at the resampling period and would otherwise have to be
smoothed by the following convolution.}

\subsubsection{FiLM Architecture Specifications}
\hl{Timestep conditioning is applied as FiLM \cite{perez2018film}. A shared MLP maps the sinusoidal code through $D \rightarrow 2D \rightarrow D$, and each residual block then applies a single affine projection $\mathrm{Linear}(D, 2C)$ whose output is split into the channel-wise scale $\gamma$ and shift $\beta$ used in Eq.~\eqref{eq:film_application}. Using one projection per block, rather than a deeper per-block MLP, keeps the conditioning pathway linear in $C$ and is the same parameterization used by DDPM \cite{ho2020denoising} and its descendants; the choice of $D = 2\max_l C_l$ ensures this projection is never rank-limiting at any level of the network.}

\subsubsection{Gaussian Noise with Cosine Schedule}
\hl{EEG is spectrally coloured, but the Gaussian forward process is a designed corruption path from the data distribution to $\mathcal{N}(0,\mathbf{I})$ rather than a model of the noise already present in the recording, and it assumes nothing about the spectrum of the signal it corrupts. Its flatness is in fact what makes the learning problem well posed: because the injected $\epsilon$ is spectrally flat while the underlying EEG is strongly non-flat, the two are statistically distinguishable, and the denoiser can predict $\epsilon$ only by learning the $1/f$ and rhythmic structure that resides in $x_0$. A coloured forward process would make the corruption spectrally similar to the signal, reducing that separability, and would additionally forfeit the closed-form marginal and reparameterization trick on which single-step training depends. The cosine schedule governs only the magnitude of noise added at each timestep, not its spectral shape. Bandpass and notch filtering attenuate, though do not eliminate, the steepest part of the $1/f$ slope below 1\,Hz.}

\subsubsection{5-Second Segments: Clinical and Computational Considerations}
\hl{The 5-second segment length was chosen on clinical grounds; it matches the timescale over which seizure morphology becomes discriminative, and it bounds the decision latency achievable in continuous monitoring. The comparison in Section~\ref{sec:window_size} shows that this choice costs nothing: a 10-second window changes accuracy by less than 0.4 percentage points and lowers AUROC from 0.9220 to 0.8713, while halving both the segments available from a recording and the rate at which decisions can be issued. The clinical and the empirical arguments therefore point the same way, which is the reason we retain 5 seconds rather than treating window length as a quantity to be tuned. A fuller characterisation across tasks is left to future work.}

\subsubsection{Bernoulli Distribution and Policy Gradient Optimization}
The Bernoulli parameterization enables policy gradient-based optimization directly for the F1-score (target metric) rather than cross-entropy (proxy metric). Stochastic sampling during training encourages exploration of decision boundaries, particularly important in imbalanced settings (6.7\% seizure prevalence) where the optimal threshold may differ from 0.5. At inference, deterministic thresholding (p > 0.5) is employed for reproducibility.}

\subsection{Binary Patient-Wise Seizure Classification}  
The binary seizure detection results (Table~\ref{tab:binary_detection}) show that the model avoids collapsing into the majority class under the severe imbalance, only 6.7\% seizure prevalence, inherent to the TUHSZ evaluation set, reaching a Weighted F1-score of 0.85 and a Macro F1-score of 0.59. \hlIII{The ablation of Section~\ref{sec:ablation_logitadj} indicates that the decision layer achieves this by relocating the operating point rather than by improving the ranking of segments, so the appropriate attribution is to the diffusion-pretrained representation together with an adjusted decision threshold, not to the reinforcement objective as a source of additional discriminative power.} Attaining 0.59 seizure recall on entirely unseen patients indicates that the learned representation transfers across the neurophysiological variability of the 279-patient cohort. \hlIII{We stress that this sensitivity is not clinically sufficient on its own: 41\% of ictal segments are missed, and closing that gap, rather than further gains in aggregate F1, is the primary obstacle to clinical use.}
\vspace{-3mm}
\subsection{Label Efficiency and Few-Shot Adaptation}
A primary objective of DiffEEG is to address the "critical gap" of label scarcity in clinical EEG monitoring. 
The k-shot analysis (Figure 4) reveals that the model effectively leverages foundational representations even with minimal supervision.
While the frozen backbone strategy provides a reliable baseline, the unfrozen (partial fine-tuning) strategy consistently yielded superior results, reaching a Weighted F1-score of 0.6578 at the 342-shot level.
This performance trajectory suggests that while self-supervised pretraining on vast amounts of unlabeled data captures generic brain activity structures, the fine-tuning of the encoder is essential to capture the nuanced clinical features required for granular subtype classification.

\vspace{-2mm}
\subsection{Interpretation of Benchmarking Results}

This section interprets the benchmarking results reported in Section~\ref{sec:benchmarking}, situating DiffEEG relative to state-of-the-art (SOTA) supervised clinical models and large-scale EEG foundation models. 
\subsubsection {Seizure Subtype Classification Benchmarking}
Our benchmarking of DiffEEG for seizure 5-subtypes classification demonstrates a critical balance between maximum discriminatory capacity and patient-independent robustness. By employing two distinct evaluation protocols, we highlight the model's ability to learn highly discriminative neural representations while contending with significant neurophysiological variability. In segment-wise experiment (Table~\ref{tab:segment_wise_results}), which serves as a measure of the model's theoretical upper bound for feature discrimination, DiffEEG reaches an Accuracy of 97.66\% and a Weighted F1-score of 97.67\%, above every result surveyed in that table. \hlIII{The split is stratified at the segment level, so segments from the same patient appear in both partitions, and within-patient temporal correlation inflates the result; the compared studies do not all adopt that protocol. The number should be read as an upper bound on what the representation can separate when patient identity is not held out, and the patient-wise results of Table~\ref{tab:patient_wise_results_updated} as the estimate that generalises.}

While the patient-wise (LOFO) generalization protocol (Table~\ref{tab:patient_wise_results_updated}) presents a more rigorous challenge for cross-subject inference, our results must be contextualized by the conditions under which each entry was obtained. Although supervised architectures such as the MBMD Transformer (2024) and Wavelet2Vec (2023) report higher average F1-scores, those benchmarks were constrained to a limited subset of only 68 patients.

\hlIII{Three differences make the entries of Table~\ref{tab:patient_wise_results_updated} only partially commensurable, and we state them rather than presenting the column as a like-for-like ranking. First, the evaluation cohorts differ by a factor of four: the higher-scoring entries evaluate 68 patients of TUHSZ v1.5.2, whereas our results span all 279 patients of v2.0.2, a larger and more heterogeneous population in which cross-patient variance is correspondingly greater. The spread across our own folds, from 0.4788 to 0.6738 weighted F1, is itself evidence that cohort composition dominates at this scale, and a 68-patient subset can fall anywhere within that range. Second, the analysis configurations differ: segment length, montage, preprocessing and class groupings are not held fixed across these studies, and our 5-second windows are shorter than those used by most of the compared methods, which reduces the context available per decision while increasing temporal resolution. Third, and most importantly, every baseline entry in this table is the value reported in its own publication; none was re-run by us. These entries therefore locate DiffEEG within the published literature but do not isolate the contribution of the model, which is precisely why we additionally report the controlled re-evaluation of Section~\ref{sec:external_probe}, where every model is run by us under one protocol.}

In contrast, DiffEEG benchmarks (Table~\ref{tab:patient_wise_results_updated}) reflect performance across a significantly larger and more diverse cohort of 279 patients, achieving a mean Weighted F1-score of 60.0\%, obtained with a 9.6M-parameter pre-trained backbone. \hlIII{The entries of Table~\ref{tab:patient_wise_results_updated} are task-trained architectures rather than pre-trained foundation models, so we draw no parameter comparison against them here. Figure~\ref{fig:model_sizes} instead places DiffEEG within the range of published EEG foundation models, and Section~\ref{sec:external_probe_discussion} weighs that trade-off where those models are re-evaluated under our own protocol.}
\subsubsection{Binary Seizure Detection (Patient-Wise)}
The binary detection task evaluates the ability to distinguish ictal segments from interictal background activity. On the official TUSZ evaluation set DiffEEG achieves an ROC-AUC of 0.7557 and an accuracy of 81.0\%, and despite the extreme class imbalance of that set (6.7\% seizure prevalence) it maintains a weighted F1-score of 85.0\% and a macro F1-score of 59\% (Table~\ref{tab:binary_detection}). \hlIII{We deliberately do not tabulate a cross-paper comparison for this task. The published binary-detection results are reported under materially different conditions: clip lengths of 12 to 16 seconds against our 5 seconds, differing patient cohorts, and in several cases metrics computed on different label definitions, and none of those models was re-run by us, so a side-by-side column would suggest a precision the underlying numbers do not support. We report instead the controlled comparison of Section~\ref{sec:adapted_tusz}, in which EEGNet, EEG-Conformer and the spatio-temporal transformer are trained by us on the same segments and evaluated beside our fine-tuned model (Table~\ref{tab:adapted_tusz}). There DiffEEG obtains the best macro F1, PR-AUC and sensitivity of the four, with disjoint confidence intervals against both transformer baselines, while the three baselines receive full end-to-end training and our model only partial unfreezing.}

\hlIII{Two observations from that table bear on how binary detection should be assessed. First, EEGNet reaches the highest accuracy of any model in this work, 0.9345, while recovering only 7.3\,\% of ictal segments; on this evaluation subset a detector that never fires scores 0.9311, so accuracy alone cannot distinguish a useful detector from a silent one, and any claim resting on it should be treated with suspicion. Second, the same table shows that EEGNet ranks segments well (AUROC 0.8803, the highest of the four) despite that failure, which localises its deficiency in the placement of the operating point rather than in the representation. The distinction between ranking quality and operating point recurs throughout our results, in the decision-layer ablation of Section~\ref{sec:ablation_logitadj} and in the window-length comparison of Section~\ref{sec:window_size}, and it is the single most consistent finding of this study.}

\begin{comment}
\begin{table}[t]
\centering
\caption{Few-Shot Seizure 5-Subtypes Classification Performance (Frozen vs. Unfrozen Backbone Strategies), 80/20 stratified split, segment-wise}
\label{tab:fewshot_comparison}
\begin{tabular}{@{}lccc@{}}
\toprule
\textbf{Mode} & \textbf{K-Shots} & \textbf{Accuracy} & \textbf{F1-Score} \\
\midrule
Frozen    & 10   & 0.3442 & 0.3604 \\
Frozen    & 50   & 0.3474 & 0.3433 \\
Frozen    & 100  & 0.5099 & 0.3690 \\
Frozen    & 200  & 0.5051 & 0.3868 \\
Frozen    & 250  & 0.5060 & 0.3760 \\
Frozen    & 342  & 0.5342 & 0.5075 \\
\midrule
Unfrozen  & 10   & 0.4496 & 0.4287 \\
Unfrozen  & 50   & 0.5053 & 0.3642 \\
Unfrozen  & 100  & 0.5080 & 0.3655 \\
Unfrozen  & 200  & 0.5446 & 0.4987 \\
Unfrozen  & 250  & 0.5576 & 0.5251 \\
Unfrozen  & \textbf{342} & \textbf{0.6564} & \textbf{0.6578} \\
\bottomrule
\end{tabular}
\end{table}
\end{comment}

\subsubsection{\hlIII{External Validation by Linear Probing on Bonn and Siena: What It Establishes}}
\label{sec:external_probe_discussion}
\hlIII{Read together, the two
corpora address the concern that our benchmarking rests on results transferred between
papers. Both are external to DiffEEG, whose pre-training saw neither corpus, and both are
evaluated here under one loader, one set of splits and one classifier family, so the only
quantity that varies between rows is the pre-trained representation. For the baselines,
externality was established by audit rather than assumption: Bonn is absent from every
model's pre-training data, while Siena is present in LaBraM's and EEGMamba's, which the
tables mark and the reading below takes into account. On the corpus with enough evaluation data to resolve
differences, DiffEEG separates significantly from two established foundation models,
BIOT and LaBraM, which indicates that the diffusion objective yields representations
transferable to seizure corpora it has never seen. Both of those models were pre-trained on
substantially more diverse data than DiffEEG: BIOT's released checkpoint aggregates six
corpora and LaBraM's approximately twenty, against the single non-seizure corpus used
here, so on this task breadth of pre-training data did not translate into better
transfer. That result should be read alongside its cost and its converse: DiffEEG carries
more parameters than either model it beats (9.67M against 3.2M and 5.8M;
Figure~\ref{fig:model_sizes}), and two further
baselines, EEGDM and CBraMod, rank segments significantly better than it does while
CBraMod does so with roughly half its parameters. Those two were also pre-trained on a
single corpus each, so the corpus count in
Tables~\ref{tab:probe_bonn} and~\ref{tab:probe_siena} does not order the models either;
what it shows is that a single, well-matched pre-training corpus is sufficient to compete
here, not that diversity is harmful. The advantage we can claim is therefore in where the operating point is
placed, not in the ordering the representation induces. That DiffEEG performs relatively better here than on
abnormality detection is consistent with its pre-training corpus: the representation was
learned from TUHSZ, and both Bonn and Siena are seizure-detection tasks, whereas
abnormality detection is a broader and less closely aligned objective.}

\hlIII{Three limitations bound these conclusions. First, and most importantly, linear
probing measures the linear separability of a frozen representation; it is not the
configuration this paper proposes. The results reported in Sections~\ref{sec:tuab}
and earlier are obtained by fine-tuning the backbone jointly with the reinforced decision
layer, and the probe deliberately removes both. These numbers should therefore be read as
a lower bound on what each architecture achieves after adaptation, and not as competing
estimates of the fine-tuned performance. \hlIII{For our own model that gap is now
measured rather than assumed: on the identical evaluation subset of
Table~\ref{tab:diffeeg_ft_gain}, fine-tuning raises PR-AUC by 0.191 and macro F1 by
0.064 over the frozen probe. Since the baselines were not fine-tuned, the corresponding
gains for them are unknown, and the probe comparison should not be taken to predict how
the ordering would change if every model were adapted.} Second, the pattern of the Siena comparison
against EEGDM is itself informative: the two models are statistically tied on the
threshold-dependent metrics but EEGDM ranks segments better, which localises the residual
gap in the ordering of the representation rather than in the placement of the operating
point, precisely the quantity the decision layer is designed to adjust. Third, the two
corpora differ in what they can support: Bonn is single-channel and close to saturated,
so it establishes competitiveness rather than any ordering, and only Siena carries enough
evaluation data at a realistic prevalence to make significance claims.}

\subsection{\hlIII{Ablation Studies}}
\subsubsection{\hlIII{Diffusion Pre-training versus Random Initialisation}}
\label{sec:ablation_pretrain}
\begin{figure}[H]
 \centering
    \includegraphics[width=1.0\linewidth]{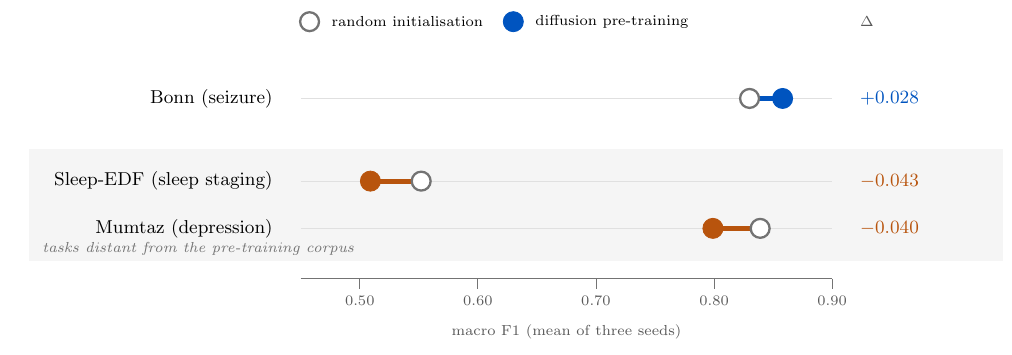}

\caption{\hlIII{Diffusion pre-training against random initialisation of the same
architecture, fine-tuned and evaluated within each corpus (macro F1, mean of three
seeds) on Bonn~\cite{andrzejak2001indications},
Sleep-EDF~\cite{kemp2000analysis} and Mumtaz~\cite{mumtaz2017electroencephalogram}. Each segment joins the two initialisations for one task; the filled marker is the
pre-trained arm and the open marker the random one. Segment colour encodes direction:
blue where pre-training is ahead, orange where random initialisation is. The benefit is
confined to the seizure task, and the sign of $\Delta$ reverses on both tasks distant
from the pre-training corpus. Seed-to-seed spread is not available, so segment length
should be read as a point estimate rather than a resolved difference.}}
\label{fig:ablation_pretrain}
\end{figure}

\hlIII{This test studies whether the pretraining of the diffusion model helped it learn something transferable rather than merely supplying a large parameter count for fine-tuning to exploit. We therefore isolate it directly: the same
architecture is fine-tuned twice on each corpus, once from the diffusion-pre-trained
weights and once from random initialisation, with everything else held fixed. Training is
within-dataset in both arms, and each configuration is run with three seeds; we report
macro F1 so that minority classes are not hidden by an imbalanced majority.
Figure~\ref{fig:ablation_pretrain} reports the outcome.}

\hlIII{The result is a qualified confirmation. On Bonn, the seizure task closest to what the model was pre-trained on, the
pre-trained initialisation is ahead by 0.028 macro F1. On sleep staging and on depression
classification it is behind, by 0.043 and 0.040 respectively, so on those corpora the
diffusion weights are not merely unhelpful but a slightly worse starting point than random
weights.}

\hlIII{We read this as evidence that the pre-training contributes \emph{within its own
domain} rather than universally. That is consistent with what the corpus contains: the
model was pre-trained on the non-seizure portion of TUHSZ, a clinical epilepsy-monitoring
corpus, and the tasks it helps on are the ones sharing that recording context and that
class of pathology. It is also consistent with the scoping we adopt throughout this paper,
where we claim transfer to seizure-related corpora and decline to claim general-purpose
EEG representation quality. An ablation that had shown uniform gains across sleep, mood
and motor imagery would have supported a broader claim than our external validation
does; this one does not, and we prefer to report it that way.}

\hlIII{Three limits bound the conclusion. The margins are small and each arm is the mean
of three seeds, whose spread we do not report here, so the ordering within any single row
should not be treated as established; what carries the finding is that the sign of
$\Delta$ is consistent across the two distant tasks and opposite on the seizure task.
Each domain is represented by one corpus, so corpus-specific effects cannot be separated
from domain-specific ones. And the comparison is within-dataset fine-tuning, which is a
different protocol from the frozen probing of Section~\ref{sec:external_probe}; it
measures what the initialisation is worth when adaptation is permitted, not the linear
separability of the representation itself.}

\subsubsection{\hlIII{Dynamic versus Fixed Logit Adjustment}}
\label{sec:ablation_logitadj}
\hlIII{The reinforced decision layer adds a scalar $a$ to the seizure logit and
subtracts it from the normal logit. Because this adjustment is antisymmetric, the
predicted seizure probability reduces to}
\begin{equation}
p \;=\; \sigma\!\left(l_1 - l_0 + 2a\right),
\label{eq:rdl_collapse}
\end{equation}
\hlIII{so a \textit{constant} $a$ translates every sample along the same axis and is
therefore equivalent to relocating the decision threshold. Such an adjustment cannot
change the relative ordering of segments, and ROC-AUC and PR-AUC, which depend only on that ordering, are provably invariant to it. Differences in these two metrics
consequently isolate the \textit{input-dependent} component of the policy, which is the
quantity the design is meant to contribute. We use this property to decompose the
decision layer into its parts.}

\hlIII{Four configurations were trained, differing only in how $a$ is produced: no
adjustment ($a=0$, the decision layer bypassed); a single learned scalar shared by all
samples; the dynamic policy $a=\pi_\phi(z)$ trained by weighted cross-entropy alone; and
the full method, $a=\pi_\phi(z)$ with the policy-gradient term. All arms share the same
pretrained backbone, partial-unfreezing scheme, probe timesteps, optimizer schedule and
class weights, and are evaluated on the same patient-disjoint evaluation cohort at a
fixed threshold of $0.5$. Each configuration was run with three random seeds; we report
mean $\pm$ standard deviation. Because the comparison is relative, the arms were trained
on a fixed subsample of 120{,}000 segments drawn from the patient-wise training split for
20 epochs, identical across all arms; absolute values are therefore slightly below those
of Table~\ref{tab:binary_detection} while the ordering between arms is directly comparable.}

\begin{table}[t]
\centering
\caption{\hlIII{Decomposition of the reinforced decision layer on patient-wise binary
seizure detection. F1$_{\text{seiz}}$ is the F1-score of the seizure class alone; the
weighted and macro variants match the metrics reported elsewhere in this work. The fixed
scalar arm reproduces the no-adjustment ROC-AUC to four decimal places, as
Eq.~\eqref{eq:rdl_collapse} requires.}}
\label{tab:ablation_logitadj}
\resizebox{\columnwidth}{!}{%
\begin{tabular}{llrccccc}
\toprule
Adjustment $a$ & RL & Params & ROC-AUC & PR-AUC & F1$_{\text{seiz}}$ & F1$_{\text{weighted}}$ & F1$_{\text{macro}}$ \\
\midrule
None ($a=0$) & -- & 0 & 0.796$\pm$0.001 & 0.297$\pm$0.001 & 0.279$\pm$0.008 & 0.822$\pm$0.011 & 0.570$\pm$0.010 \\
Fixed scalar & -- & 1 & 0.796 & 0.295 & 0.281 & 0.823 & 0.571 \\
Dynamic $\pi_\phi(z)$ & -- & 1.44M & 0.792$\pm$0.002 & 0.288$\pm$0.003 & 0.279$\pm$0.011 & 0.823$\pm$0.016 & 0.571$\pm$0.013 \\
Dynamic $\pi_\phi(z)$ & \checkmark & 1.44M & 0.792$\pm$0.001 & 0.284$\pm$0.010 & 0.281$\pm$0.009 & 0.826$\pm$0.013 & 0.573$\pm$0.011 \\
\bottomrule
\end{tabular}}
\end{table}

\hlIII{Table~\ref{tab:ablation_logitadj} supports three observations. First, the fixed
scalar arm matches the no-adjustment arm on ROC-AUC to within $+0.0000$, confirming
empirically the invariance argued above and validating the implementation. Second, making
the adjustment input-dependent does not improve discrimination: the dynamic policy changes
ROC-AUC by $-0.0037$ relative to no adjustment, a small but consistent degradation of
roughly twice the seed-to-seed standard deviation, obtained at a cost of 1.44M additional
parameters. Third, the policy-gradient term contributes a further $-0.0005$, which is
indistinguishable from zero.}

\hlIII{The direction of these changes is itself informative. Across the four arms ROC-AUC
decreases monotonically while F1 increases monotonically, and every F1 difference
($+0.0017$ to $+0.0021$ weighted) lies within the seed-to-seed spread of $\pm0.011$ to
$\pm0.016$. This pattern is the signature of a threshold shift rather than improved
ranking: the decision layer relocates the operating point in a direction that favours the
minority class at a fixed threshold of $0.5$, without ordering segments any better. The
fixed scalar arm shows that a single parameter achieves the same effect. We therefore do
not claim that the reinforced decision layer improves discrimination; its measurable
contribution under this protocol is equivalent to learning an operating point, and we
report this negative result so that the reported performance can be attributed correctly.
We note that these runs use a reduced training budget, and that a full-budget replication
would be required before concluding that the same holds at the scale of
Table~\ref{tab:binary_detection}.}

\subsubsection{\hl{Effect of Analysis-Window Length}}
\label{sec:window_size}
\hl{The 5-second analysis window is a design choice rather than a tuned parameter, so we
tested it directly. We repeated the TUAB abnormality-detection protocol above with
10-second windows, holding the pretrained backbone, the fine-tuning schedule, the probe
timesteps and the partial-unfreezing scheme fixed, so that segment length is the only
variable. Table~\ref{tab:window_size} reports both configurations on the same held-out
evaluation set.}

\begin{table}[H]
\centering
\caption{\hl{Effect of analysis-window length on TUAB abnormality detection.}
\hlIII{All configurations share the pre-trained backbone, partial-unfreezing scheme, probe
timesteps, optimizer schedule and class weighting. The first two columns isolate window
length with the reinforced decision layer active in both, and the third repeats the
10\,s configuration with the policy-gradient term disabled ($\lambda=0$) so that the
contribution of the reinforcement objective can be read off directly.}
\hl{Metrics are computed on the held-out TUAB evaluation set.}}
\label{tab:window_size}
\begin{tabular}{lccc}
\toprule
\textbf{Metric} & \textbf{5\,s (RL)} & \textbf{10\,s (RL)} & \textbf{10\,s (no RL)} \\
\midrule
Accuracy                 & 80.00\%        & 79.00\%   & \textbf{80.00\%} \\
AUROC                    & \textbf{0.92} & 0.87   & 0.87 \\
PR-AUC                   & 0.86         & \textbf{0.87} & 0.87 \\
Sensitivity              & --             & \textbf{0.79} & 0.78 \\
Specificity              & --             & 0.80   & \textbf{0.81} \\
\midrule
Training segments        & 735{,}414      & 372{,}510 & 372{,}510 \\
Evaluation segments      & 72{,}291       & 36{,}945  & 36{,}945 \\
Decision latency         & 5\,s           & 10\,s     & 10\,s \\
\bottomrule
\end{tabular}
\end{table}

\hl{Doubling the window leaves accuracy essentially unchanged, within 0.4 percentage
points, while halving both the number of segments obtainable from a recording and the rate
at which decisions can be issued.} \hlIII{With the decision layer active in both
configurations, the 5\,s setting retains a clear advantage in AUROC (0.9220 against
0.8713) while the 10\,s setting is marginally ahead on PR-AUC (0.8715 against 0.8600), so
doubling the window does not improve the ranking of segments and costs decision latency.}

\hlIII{The third column isolates the reinforcement objective at fixed window length, and
its effect is negligible: disabling the policy-gradient term changes AUROC by $0.0001$
(0.8713 against 0.8714) while shifting the operating point, raising specificity by
0.0133 and lowering sensitivity by 0.0069. This independently reproduces, on a different
corpus and a different window length, the conclusion of the ablation in
Section~\ref{sec:ablation_logitadj}: the reinforced decision layer relocates the operating
point rather than improving discrimination. It also confirms that the AUROC difference
between the two window lengths is a property of the window rather than of the
reinforcement term.} \hl{On the evidence available, longer windows bring no measurable
benefit on this task, which supports retaining 5-second segments for continuous
monitoring.}

\subsubsection{\hlIII{Clean versus Noised Multi-Timestep Probing}}
\label{sec:ablation_probe_noise}
\hlIII{Multi-timestep probing passes the unmodified segment to the encoder and varies only
the timestep index, whereas the diffusion literature evaluates a denoiser on inputs
corrupted to the level that index denotes. Because the choice is not self-evident, we
measured it. We built a second descriptor, identical in every respect except that each
segment was corrupted before extraction by the pre-training forward process,
$\mathbf{x}_t = \sqrt{\bar{\alpha}_t}\,\mathbf{x}_0 + \sqrt{1-\bar{\alpha}_t}\,\epsilon$
with $\epsilon \sim \mathcal{N}(0,\mathbf{I})$, using the same cosine schedule and a fixed
random seed. Both descriptors are $2{,}400$-dimensional, come from the same frozen
backbone and the same probe timesteps, and were evaluated under the same frozen
linear-probe protocol applied to the external corpora above, on the TUHSZ subset defined
in Section~\ref{sec:adapted_tusz}: binary seizure detection, 100{,}000 training and
50{,}000 held-out segments at the natural prevalence of about 6.7\,\% seizure, a ten-point
regularization sweep selected on a stratified validation split, and 1{,}000 bootstrap
resamples of the evaluation set. Panel A of Table~\ref{tab:ablation_probe_noise}
reports the comparison.}

\begin{table}[H]
\centering
\caption{\hlIII{Clean versus noised multi-timestep probing. Panel A is the
protocol-matched comparison: frozen backbone, frozen linear probe, TUSZ binary detection
on 50{,}000 held-out segments, regularization selected on a validation split, 95\%
confidence intervals from 1{,}000 bootstrap resamples. Every metric separates with
disjoint intervals. Panel B is a supplementary diagnostic at fixed regularization
($C=1$) on a 30{,}000-segment training subsample, and is therefore not directly
comparable to Panel A in absolute terms; it isolates each probe timestep in turn.
$\sqrt{1-\bar{\alpha}_t}$ is the noise magnitude that index carries under the
pre-training schedule.}}
\label{tab:ablation_probe_noise}
\begin{tabular}{lccc}
\multicolumn{4}{l}{\textbf{\hlIII{Panel A: protocol-matched, all five timesteps}}} \\
\toprule
\textbf{Metric} & \textbf{Clean input} & \textbf{Noised input} & \textbf{$\Delta$} \\
\midrule
AUROC        & \textbf{0.8129} & 0.7475 & $+0.0654$ \\
PR-AUC       & \textbf{0.3005} & 0.2165 & $+0.0840$ \\
Accuracy     & \textbf{0.7598} & 0.7290 & $+0.0307$ \\
Weighted F1  & \textbf{0.8166} & 0.7943 & $+0.0223$ \\
Macro F1     & \textbf{0.5748} & 0.5411 & $+0.0337$ \\
Sensitivity  & \textbf{0.7276} & 0.6468 & $+0.0807$ \\
Specificity  & \textbf{0.7622} & 0.7351 & $+0.0270$ \\
\bottomrule
\end{tabular}

\vspace{2mm}
\begin{tabular}{lccc}
\multicolumn{4}{l}{\textbf{\hlIII{Panel B: single timestep, fixed $C=1$, 30k subsample}}} \\
\toprule
\textbf{Probe timestep} & \textbf{$\sqrt{1-\bar{\alpha}_t}$} & \textbf{AUROC (clean)} & \textbf{AUROC (noised)} \\
\midrule
$t=50$   & 0.091 & 0.8142 & 0.7465 \\
$t=250$  & 0.393 & 0.8146 & 0.5763 \\
$t=500$  & 0.713 & 0.8148 & 0.5240 \\
$t=750$  & 0.926 & 0.8149 & 0.5035 \\
$t=950$  & 0.997 & 0.8153 & 0.5014 \\
\midrule
All five & --    & 0.8174 & -- \\
\bottomrule
\end{tabular}
\end{table}

\hlIII{Passing the clean segment is the better choice, and the margin is not marginal:
every one of the seven metrics separates with disjoint 95\% confidence intervals, and the
two metrics that matter most under 6.7\% prevalence move furthest, AUROC by 0.0654 and
PR-AUC by 0.0840. The mechanism is visible in Panel B. Corrupting the input destroys the
seizure-discriminative content in proportion to the noise magnitude the index carries:
at $t=50$, where $\sqrt{1-\bar{\alpha}_t}=0.091$ and the segment is still largely intact,
the noised probe retains an AUROC of 0.7465, but by $t=950$, where the input is
essentially white noise, it reaches 0.5014, which is chance. A denoiser evaluated on
$\mathbf{x}_t$ can only report what survives the corruption, and past the middle of the
schedule almost nothing does. The two panels should not be compared entry by entry: Panel
B trains on 30{,}000 segments at a fixed $C=1$, so its noised five-timestep counterpart is
not the 0.7475 of Panel A, and the near-equality between that value and the $t=50$ entry
of Panel B is an artefact of the differing budgets rather than a finding.}

\hlIII{Panel B also revises a claim we made in earlier drafts. We had motivated
multi-timestep probing on the premise that the encoder's representations vary meaningfully
across the noise schedule, so that concatenating five of them captures complementary views
of the signal. With clean inputs this premise is not supported. The five single-timestep
probes span 0.0011 in AUROC (0.8142 to 0.8153), the concatenation of all five improves on
the best single block by 0.0021, and the pooled block activations correlate above 0.995
across every pair of timesteps. What varies across $\mathcal{T}$ is the FiLM modulation
the timestep embedding applies, and its effect on the pooled encoder features is small
relative to the effect of the signal itself. The concatenation is therefore best read as a
mild ensembling gain over a largely redundant set of views, not as the aggregation of
distinct representational scales. We retain the five-timestep descriptor because it is
what the reported results were produced with and because the gain, while small, is
consistent, but a single-timestep descriptor at one fifth the dimensionality would cost
approximately 0.002 AUROC on this task.}

\section{Conclusion}
\label{sec:conclusion}

This paper proposes DiffEEG, a 9.6M-parameter self-supervised foundation model that combines denoising diffusion pre-training with a reinforced decision layer (RDL) to address the dual challenges of label scarcity and extreme class imbalance in clinical EEG analysis.

\hl{\textbf{Technical Contributions:} The architectural choices in DiffEEG are stated explicitly and grounded either in the structure of the network or in established diffusion-model practice. The channel expansion $[1,2,4,8]$ enables hierarchical multi-scale feature learning suited to EEG's multi-frequency structure. The embedding width $D=512$ follows the convention $D = 2\max_l C_l$, which makes the deepest FiLM projection square (Eq.~\eqref{eq:embed_dim}); it was aligned to the channel schedule rather than selected by search, and smaller widths were not evaluated. The bottleneck operates at the $L/8$ resolution implied by the multiplier schedule, with three residual blocks giving the self-attention layers sufficient depth within a single-A100 memory budget. Upsampling by linear interpolation followed by a learned convolution avoids the checkerboard artefacts of strided transposed convolution, which is undesirable when the model output is itself a waveform. FiLM conditioning applies a shared $D \rightarrow 2D \rightarrow D$ timestep MLP and a single per-block $\mathrm{Linear}(D, 2C)$ projection, following DDPM \cite{ho2020denoising} and Perez et al. \cite{perez2018film}. The cosine variance schedule suits EEG's spectrally coloured noise, and a fixed $T=1000$ schedule provides comparability with the diffusion literature and reproducibility across runs. Bernoulli parameterization enables policy-gradient optimization directly against the target F1-score metric. The 5-second segment length reflects the timescale of seizure morphology while remaining computationally practical for continuous monitoring.}

\hl{\textbf{Clinical Performance on Realistic Imbalanced Data:} Under clinically rigorous evaluation (patient-wise LOFO, 279 patients, 6.7\% seizure prevalence), DiffEEG achieves 59\% seizure recall with 85\% weighted F1 for binary detection, retaining usable sensitivity despite extreme class imbalance. \hlIII{This level is nonetheless well short of clinical adequacy: at 59\% recall, 41\% of ictal segments are missed, whereas continuous monitoring is generally expected to operate at high sensitivity with alarms adjudicated by a reviewer. We therefore position these results as evidence that the representation transfers under severe imbalance, not as a claim of deployment readiness.} For seizure subtype classification, mean performance of 59\% F1 across diverse patients indicates partial cross-patient robustness, with best-fold performance (67.4\%) suggesting potential for patient-tailored fine-tuning.}

\hl{\textbf{Cross-Dataset Generalization:}} \hlIII{We deliberately do not claim that DiffEEG performs well on EEG tasks in general. The evidence supports a narrower and more specific statement: the representation transfers well to \emph{seizure-related} corpora. Under the controlled linear-probing protocol of Section~\ref{sec:external_probe}, DiffEEG is competitive with five established foundation models across two external seizure databases, Bonn (Table~\ref{tab:probe_bonn}) and Siena (Table~\ref{tab:probe_siena}), attaining the best accuracy, weighted F1 and specificity of every model compared on Siena, including two that were themselves pre-trained on that corpus, in addition to its results on TUHSZ. This alignment is consistent with its pre-training corpus, which consists of non-seizure TUHSZ recordings. On the broader task of abnormality detection (TUAB), which is not seizure-specific, the fine-tuned model reaches 0.922 AUROC, but we note that this is a fine-tuned result compared against published numbers rather than a controlled re-evaluation. Claims of general-purpose EEG representation quality would require the same controlled protocol applied across non-seizure paradigms, which this study does not undertake.}

\hl{\textbf{Computational Efficiency:} At 9.6M parameters, DiffEEG remains competitive with larger transformer-based foundation models at a lower memory footprint, making it practical for resource-constrained clinical environments (Figure~\ref{fig:model_sizes}).} \hlIII{We note the limit of this claim: DiffEEG is larger than two of the four baselines it is probed against in Section~\ref{sec:external_probe}, and CBraMod matches or exceeds its ranking quality on Siena with roughly half the parameters, so parameter count alone does not favour our model across the board.}

However, significant challenges remain. Cross-patient generalization shows high variance (F1: 0.48-0.67 across folds) reflecting neurophysiological diversity, electrode placement differences, and recording equipment variations. Future research directions include: (1) data-adaptive noise schedules explicitly modeling EEG spectral structure; (2) multi-site training data to improve cross-institution generalization; (3) patient-specific fine-tuning strategies leveraging foundation model representations; (4) hardware deployment optimization for point-of-care seizure monitoring.

DiffEEG demonstrates that diffusion-based self-supervised learning, combined with metric-aware reinforcement learning and rigorous architectural design validation, provides a promising direction for label-efficient seizure analysis. 

\section*{Declaration of Generative AI and AI Assisted Technologies in the Manuscript Preparation Process}
During the preparation of this work, the authors used ChatGPT and Claude to improve language and readability. After using these tools, the authors reviewed and edited the content as needed and take full responsibility for the content of the published article.
\bibliographystyle{iopart-num}
\bibliography{bibliography}

\end{document}